\RequirePackage{fix-cm}
\documentclass{svjour3}                     
\smartqed  
\usepackage{multirow}
\usepackage{cite}
\usepackage{subcaption}
\usepackage{graphicx}
\graphicspath{ {figures/} }
\usepackage[fleqn]{amsmath}
\usepackage{epstopdf}
\usepackage{mathtools}
\usepackage{float}
\usepackage{tabularx}
\usepackage{algorithm}
\usepackage{algpseudocode}
\usepackage[misc]{ifsym}
\usepackage{enumitem}
\newcommand{\mytilde}{\raise.17ex\hbox{$\scriptstyle\sim$}}
\usepackage{tikz}
\newcommand*\circled[1]{\tikz[baseline=(char.base)]{
            \node[shape=circle,draw,inner sep=1pt] (char) {#1};}}
\journalname{}
\begin{document}

\title{Indoor UAV scheduling with Restful Task Assignment Algorithm
}


\author{Yohanes Khosiawan         \and
        Izabela Nielsen 
}


\institute{Y. Khosiawan \and I. Nielsen (\Letter) \at
              Department of Materials and Production\\
              Aalborg University, Aalborg 9220, Denmark\\
              \email{\{yok, izabela\}@make.aau.dk}           
}

\date{Received: date / Accepted: date}

\maketitle

\begin{abstract}
Research in UAV scheduling has obtained an emerging interest from scientists in the optimization field. When the scheduling itself has established a strong root since the 19th century, works on UAV scheduling in indoor environment has come forth in the latest decade. Several works on scheduling UAV operations in indoor (two and three dimensional) and outdoor environments are reported. In this paper, a further study on UAV scheduling in three dimensional indoor environment is investigated. Dealing with indoor environment\textemdash where humans, UAVs, and other elements or infrastructures are likely to coexist in the same space\textemdash draws attention towards the safety of the operations. In relation to the battery level, a preserved battery level leads to safer operations, promoting the UAV to have a decent remaining power level. A methodology which consists of a heuristic approach based on Restful Task Assignment Algorithm, incorporated with Particle Swarm Optimization Algorithm, is proposed. The motivation is to preserve the battery level throughout the operations, which promotes less possibility in having failed UAVs on duty. This methodology is tested with 54 benchmark datasets stressing on 4 different aspects: geographical distance, number of tasks, number of predecessors, and slack time. The test results and their characteristics in regard to the proposed methodology are discussed and presented.
\end{abstract}

\section{Introduction}
\label{sec:sec_intro}
In the recent years, there has been a remarkable rise of applications and research interests in UAVs. Outdoor UAV operations has been applied through various domains and enabling systems such as \cite{shima2005assignment,shima2005uav,ahner2006assignment,piciarelli2013outdoor,von2015deploying}, while reported works on UAV operations for indoor environment \cite{khosiawan2016system} are still limited. However, there is an emerging demand to employ UAVs for indoor operations \cite{khosiawan2016system}.
For instance, UAVs can be utilized to perform tasks such as inspection and material handling in a manufacturing environment. To fulfill such tasks, the UAV is equipped with an imaging device and a gripper. UAVs perform the tasks at the designated positions with the required equipment. The execution manner of these tasks also considers the total makespan or/and battery consumption. To accomplish such a goal, a schedule of task executions needs to be generated, and the respective instructions to be sent to the UAVs. During the schedule generation, several constraints need to be satisfied to produce a feasible schedule.

The scheduling problem in this paper includes an optimization problem of assigning execution time of tasks in respect to the available limited resources such as UAVs and positions, whose class has been proven to be NP-hard \cite{garey1979computers}. Furthermore, the planning horizon may include non-determined recharge actions (as needed) during the operations. Unlike in \cite{khosiawan2016task}, a variable recharge time is employed in this paper. Any sufficient time range (over the defined threshold) can be used for performing a recharge. Apart from that, other actions such as hover and wait on ground may also be included to represent feasible operations. In such a problem, obtaining an optimum solution takes a non-linear computation time in regard to the problem scale. During the numerical experiments (Section \ref{sec:sec_num_exp}), benchmark datasets on both lab scale and industrial scale indoor environments are used. In the lab scale environment itself, with small numbers of tasks, hours of computation time are shown insufficient to complete the optimization process. This entails the need of a heuristic-based approach, where a near optimum (good quality feasible) solution is able to be obtained in a reasonable amount of time. Such computation time, together with the exponentially growing one obtained from experiments with IBM ILOG CPLEX, are presented in Section \ref{sec:sec_num_exp}.

In \cite{khosiawan2016task}, the authors developed a methodology which includes the earliest available time heuristic which is incorporated with a metaheuristic algorithm called Particle Swarm Optimization (PSO) to obtain a good quality feasible solution in a quick computation time. The objective in \cite{khosiawan2016task} is to find a minimized total makespan. Meanwhile in this paper, the objective of the optimization is to minimize the total energy consumption. The tasks are given time windows, from which can be inferred that as long as the time window constraints are not violated, the execution of the following business process will not be delayed. Hence, optimizing battery consumption become relevant to keep the operational cost minimum, while inducing short flight paths simultaneously. This objective is associated with the awareness of keeping the battery level preserved achieved through the Restful Task Assignment Algorithm (RTAA). Battery is the source of life for the UAV, where battery failure may lead to the failure of the task and the UAV itself. In addition, the map of the operational environment is constructed with directed paths, where the layout is designed to support collision avoidance throughout the operations (described in Section \ref{sec:sec_problem}). The main contributions in this work are described as follows.
\begin{enumerate}[topsep=0pt]
\item Developed a methodology for safe scheduling which includes:
\begin{itemize}
\item a novel heuristic based on RTAA for constructing a schedule from a task sequence\\
RTAA promotes the preserved battery level of the UAVs throughout the operations.
The tendency of the preserved battery level avoids UAVs from suddenly \textit{dropping out of the game} due to an unforeseen marginal remaining battery level.
\item an incorporation of RTAA-based heuristic with PSO to obtain a good quality feasible solution in a short computation time\\
PSO parameters used in Section \ref{sec:sec_num_exp} are inspired by the pilot study in \cite{khosiawan2016task}, where parameter analysis was extensively performed.
\item a directed-path three-dimensional indoor environment map\\
The tasks are to be executed in a particular three-dimensional indoor environment, which is defined in a map.
In the map, the paths are constructed in a directed manner, also in coherent with the UAV flying behavior, which altogether supports collision avoidance during the operations.
\end{itemize}
\item Performed numerical experiments and analysis on the characteristics of the resulting schedules on multiple datasets.
There are 54 benchmark datasets stressing on different aspects, i.e. geographical distance, number of tasks, number of predecessors, and slack time.
The characteristics analyzed from the resulting schedules are the total energy consumption, makespan, and computation time.
\end{enumerate}

The remainder of this paper is organized as follows. In Section \ref{sec:sec_literature}, some related works on scheduling tasks with time windows and UAV scheduling in indoor environment are reviewed. Then, in Section \ref{sec:sec_problem}, the formulation of the problem in this paper is presented. In Section \ref{sec:sec_methodology}, RTAA and its incorporation with PSO are described. Afterwards, the conducted numerical experiment is presented and analyzed in Section \ref{sec:sec_num_exp}, followed with the conclusion in Section \ref{sec:sec_conclusion}.

\section{Literature review}
\label{sec:sec_literature}
Scheduling has been a consistently demanding research interest which involves numerous scientists and specific problems in the optimization area. In 1988, a study on deciding the minimum fleet size for minimum traveling salesman problem with time windows \cite{desrosiers1988lagrangian} was conducted to be integrated into the previous related work \cite{desrosiers1984routing} in 1986. The evaluation of the optimum number of fleet at the first node of a branch-and-bound tree will reduce the number of infeasible routes and the size of the tree to be explored. This approach, wherein an optimal solution is aimed at, is suitable for scheduling tasks significantly prior to the execution where changes towards the schedule during the execution are less likely to happen, e.g. in school bus transportation problem. When different numerous tasks are performed by multiple agents on a daily basis, where the tasks may come until right before the operation is started, a non-exhaustive near optimal search within a short time becomes more practical.

A work on optimization using column generation approach for the vehicle routing problem with time windows \cite{desrochers1992new} was carried out in 1992, where four essential rules for reducing the time window width were introduced as a part of the methodology. Rules 2 and 3 can be applied when there is only a single agent or there is no position occupation during the task execution. For instance, when a time window of a predecessor is trimmed upfront because the successor cannot be started sooner anyway (regardless of the predecessor's sooner completion time), the required position to be accessed may be occupied by another agent later. Executing the predecessor earlier (and the successor is not immediately executed afterwards) may become its only option to satisfy the given time window constraint. In regard to rule 1 and 4, the respective time window examples in \cite{desrochers1992new} which may lead to some unfeasible task assignments is not the focus of this paper, and hence such a condition is not reflected in the benchmark datasets (they are discussed later in Section \ref{sec:sec_num_exp}).

In the following years, heuristic-based approach is viewed to remain as a viable alternative for solving large-scale optimization problem in a short time \cite{desrosiers1995time}. For scheduling problem with time windows, a study on backtracking techniques for the job shop scheduling constraint satisfaction problem was conducted. To come up with a feasible solution as fast as possible, one essential remark to keep in mind is the expensive computation when a large complex conflict is met, for proving that some particular assignments need to be undone.

The beginning of aircraft scheduling problem \cite{desaulniers1997daily} was mainly flagged in 1997, using branch-and-bound approach supported by column generation technique. The goal is to gain a substantial profit improvement over the existing airline's solution. The nature of such problem allows a longer computation time which is deemed reasonable. Recently, the application of aircraft has been widely explored, especially Unmanned Aerial Vehicle (UAV) into various application domains in both outdoor and indoor environments \cite{khosiawan2016system}. It can be deduced that a system of UAV application mainly comprises environment (mapping and positioning system), Unmanned Aerial System (UAV, ground station, control module), and task (scheduling). Those elements are affecting while enabling one another simultaneously. One problem instance of such kind involves numerous tasks and constraints to be considered in the scheduling process. This brings the appeal of heuristic-based approach being able to solve the problem with a good quality feasible solution in a reasonable time. A modified receding horizon task assignment heuristic which focuses on scheduling split jobs of long surveillance missions to maintain a persistent UAV operations was conducted \cite{kim2014concerted}, where an indoor system prototype was tested in \cite{song2013persistent}. Another heuristic-based is also used for UAV scheduling in two-dimensional environment with no constraint \cite{semiz2015task}. Under the metaheuristic family, PSO has been investigated for job shop and flexible manufacturing (where the employment of unmanned vehicles is found) scheduling problems in \cite{liu2007investigation, mousavi2017multi, khosiawan2016task}, and it is postulated to be effective and efficient. Such a metaheuristic algorithm is fast and easy to implement \cite{sorensen2013metaheuristics,liu2007investigation}. Such an approach towards multiple tasks scheduling problem for UAV operations in indoor environment was piloted in \cite{khosiawan2016task}. The objective is to produce a schedule with a minimized total makespan in a short time. The pilot work serves as the ground to identify further issues and the respective solution - which are discussed in the following sections in this paper.

\section{Problem definition}
\label{sec:sec_problem}
Numerous task executions by UAVs in indoor environment require a thorough scheduling mechanism to reduce cost (e.g. time and energy) and promote safety towards both the users (human-labors) and the system itself (e.g. UAVs, indoor infrastructure). In the previous work \cite{khosiawan2016task}, the total time of the operation is sought to be minimized to save both time and financial cost, and to allow the following business process to continue (e.g. surface inspections of the wind turbine blades need to be done before the trucks are booked for transporting them).
In this study, the goal is to produce a schedule of task executions with a minimized total energy consumption and preserve the battery level throughout the operations. Additionally, a task has a time window attribute, in which the task must be executed. Furthermore, when dealing with UAVs in indoor environment, no sudden out-of-power is expected since it may lead to human injuries or infrastructure damages. Complying to these conditions, a new heuristic-based method which minimizes the total battery consumption with the awareness of keeping the battery level preserved is required; not strictly utilizing the UAV without recharge until the battery is depleted. This is the limitation that is exposed in \cite{khosiawan2016task}, where a full recharge is done only when the battery is completely or nearly depleted that the UAV can't perform its following action.

An instance of a task dataset is depicted in Table \ref{table:tasks}. Each task data consists of a task identifier, a start position, an end position (if it is an inspection task, then the value is the same as the start position; i.e. task \circled{1},\circled{2},\circled{5},\circled{7},\circled{8}), a processing time, a release date, a due date, and a predecessor list. In this study, the benchmark datasets used in Section \ref{sec:sec_num_exp} are tasks with time windows, while a support for assigning tasks without time window is induced in the proposed methodology - discussed later in Section \ref{sec:sec_methodology}. A characteristic which is introduced by the time window attribute is slack time. Slack time indicates the room where the execution time may be shifted around within the time window. For instance, task \circled{10} in Table \ref{table:tasks} has a time window of 281 seconds from 726\mytilde1007, and a slack time of (281-44=)237 seconds\textemdash since the processing time of task \circled{10} is 44 seconds.
\begin{table} [htp]
\caption{Example of task dataset with 10 tasks}
\begin{center}
    \begin{tabularx}{\textwidth}{ X|X|X|X|X|X|X }
    \hline\noalign{\smallskip}
    Task ID & Start\newline Position & Processing\newline Time & End\newline Position & Release\newline date & Due date & Predecessor\newline (Task ID)\\
    \noalign{\smallskip}\hline\noalign{\smallskip}
    1 & c3 & c3 & 10 & 726 & 1077 & 8\\
    2 & c1 & c1 & 10 & 1007 & 1379 & 10\\
    3 & d4 & f2 & 39 & 1429 & 1745 & 7\\
    4 & e2 & b2 & 42 & 827 & 1230 & 9\\
    5 & d4 & d4 & 10 & 0 & 382 & -\\
    6 & f1 & e2 & 35 & 1428 & 1843 & 7\\
    7 & d1 & d1 & 10 & 1007 & 1429 & 9;10\\
    8 & b3 & b3 & 10 & 292 & 726 & -\\
    9 & c4 & b3 & 35 & 382 & 827 & 5\\
    10 & a2 & c2 & 44 & 726 & 1007 & 8\\
    \hline
    \end{tabularx}
\end{center}
\label{table:tasks}
\end{table}

There are two types of task: inspection and material handling tasks. They expose 3 types of actions: perform task - material handling, perform task - inspection, and fly to.
On top of that, there are 3 additional types of action that the UAV perform throughout the operations: hover, wait on ground, and recharge. Those actions are depicted in Table \ref{table:table_action_type}. The time for performing material handling is composite, where it comprises 30 seconds for loading and unloading the payload, and a variable flying time from the start to the end position. Such a level of task abstraction is needed to reduce the complexity of the computation. Throughout the operations, actions of perform task (material handling and inspection), fly to, and hover are the contributing factors in the battery consumption.
\begin{table} [htp]
\caption{Action types}
\begin{center}
    \begin{tabular}{ l c }
    \hline\noalign{\smallskip}
    Action Type & Execution time (s)\\
    \noalign{\smallskip}\hline\noalign{\smallskip}
    Perform task - material handling & 30 + \$\textit{fly to}\$\\
    Perform task - inspection & 10 \\
    Fly to & (as defined in the map)\\
    Hover & (as required)\\
    Wait on ground & (as required)\\
    Recharge & $\geq$270 \\ \hline
    \end{tabular}
\end{center}
\label{table:table_action_type}
\end{table}

\begin{figure}[H]
  \centering
  \includegraphics[width=0.9\linewidth]{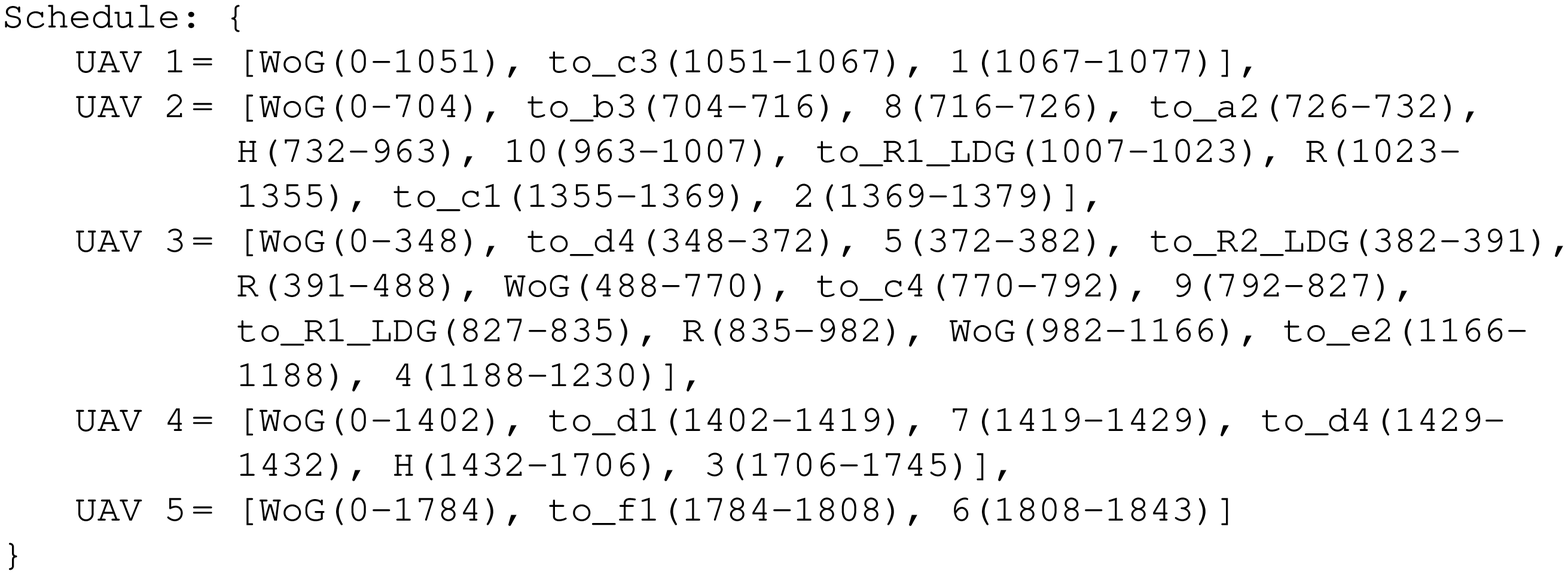}
  \caption{Representation of schedule of the UAV operations}
  \label{fig:schedule}
\end{figure}

\begin{figure}[H]
  \centering
  \includegraphics[width=0.8\linewidth]{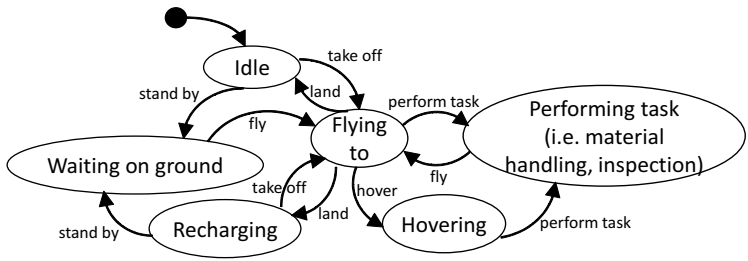}
  \caption{State machine of the UAV throughout the operations}
  \label{fig:state}
\end{figure}

Furthermore, an example of schedule which describes the solution representation for the addressed problem is depicted in Figure \ref{fig:schedule}.
It corresponds to the task dataset in the aforementioned Table \ref{table:tasks}.
A schedule consists of the assigned tasks and the other actions, whose composition conforms to the UAV state machine depicted in Figure \ref{fig:state}.
Upon the completion of its last task, the UAV goes back to a nearest recharge station by default - such an action is not included in the schedule, while the assurance of having enough battery to go back is taken into account - the imposing mechanism is explained later in Section \ref{sec:subsec_rtaa}.

\begin{figure}
  \centering
  \begin{subfigure}{0.8\textwidth}
  \includegraphics[width=\linewidth]{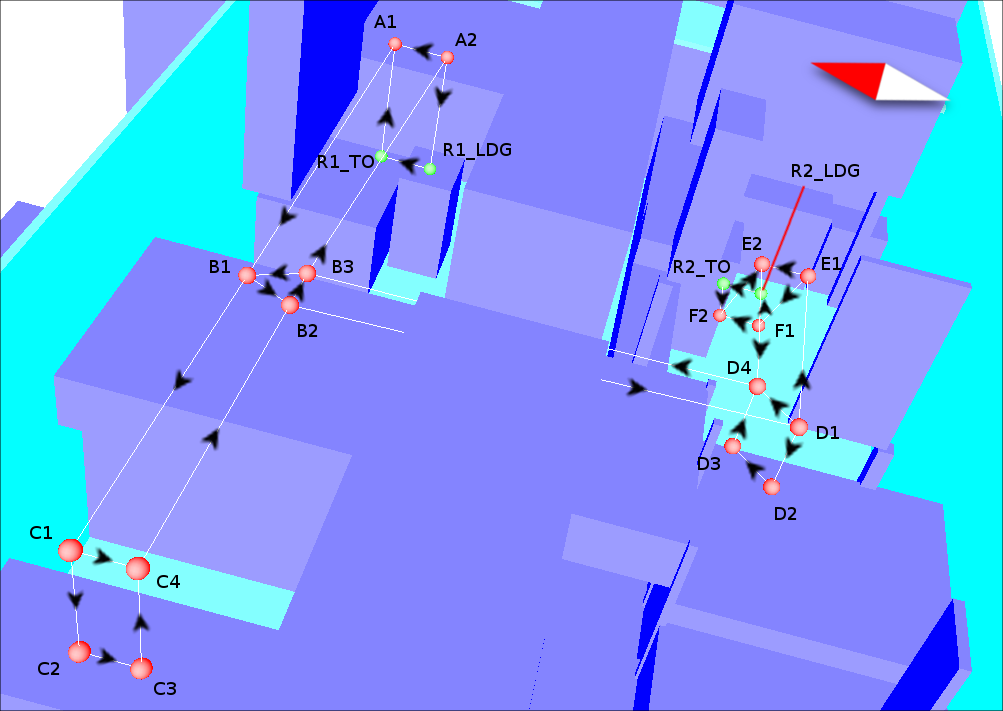}
  \caption{Positioning-system-based representation}
  \end{subfigure}
  \begin{subfigure}{0.8\textwidth}
  \includegraphics[width=\linewidth]{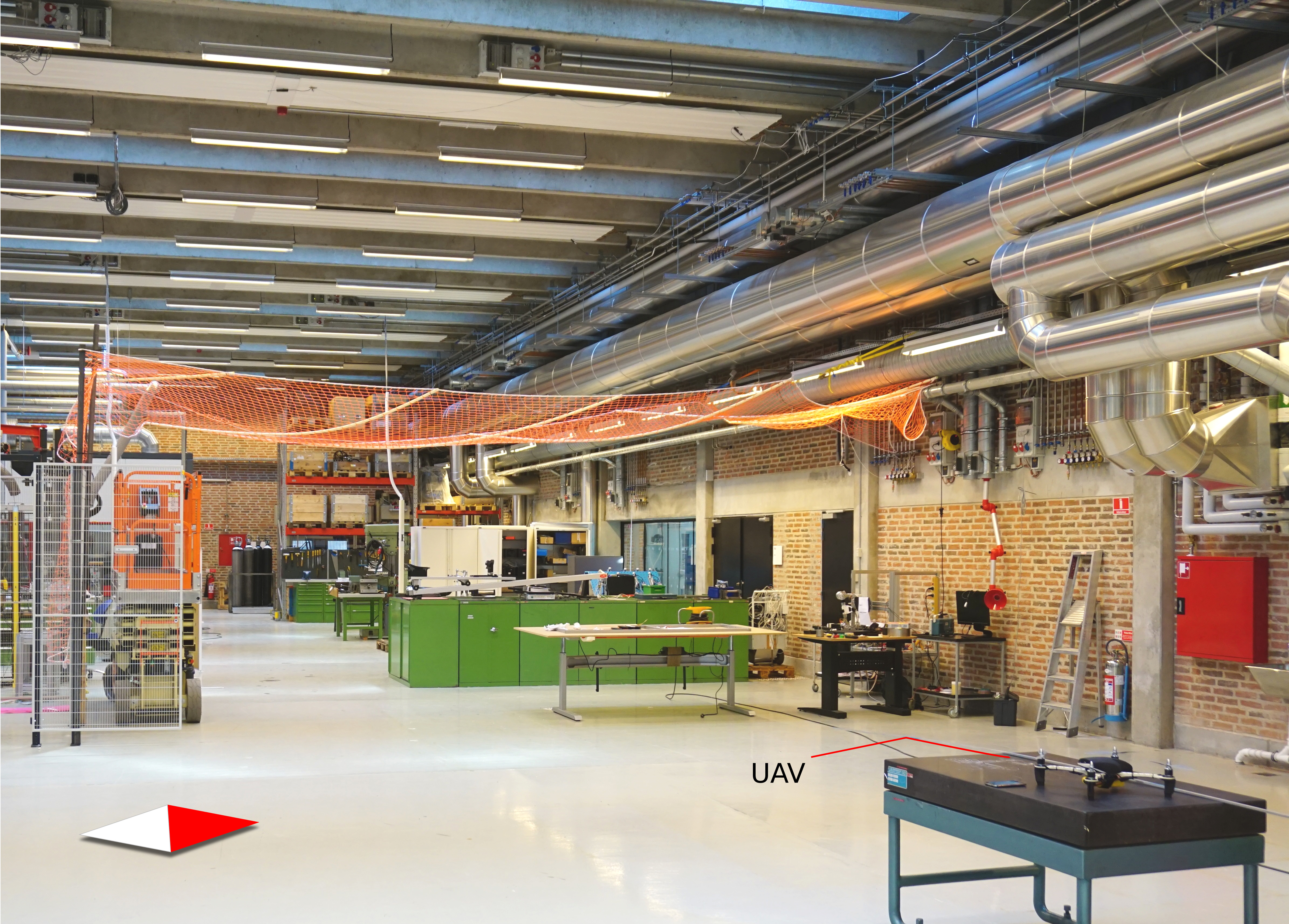}
  \caption{Real-world representation}
  \end{subfigure}
  \caption{Lab scale indoor environment for the UAV operations}\label{fig:env}
\end{figure}

\begin{figure}[H]
  \centering
  \begin{subfigure}{0.4\textwidth}
  \includegraphics[width=\linewidth]{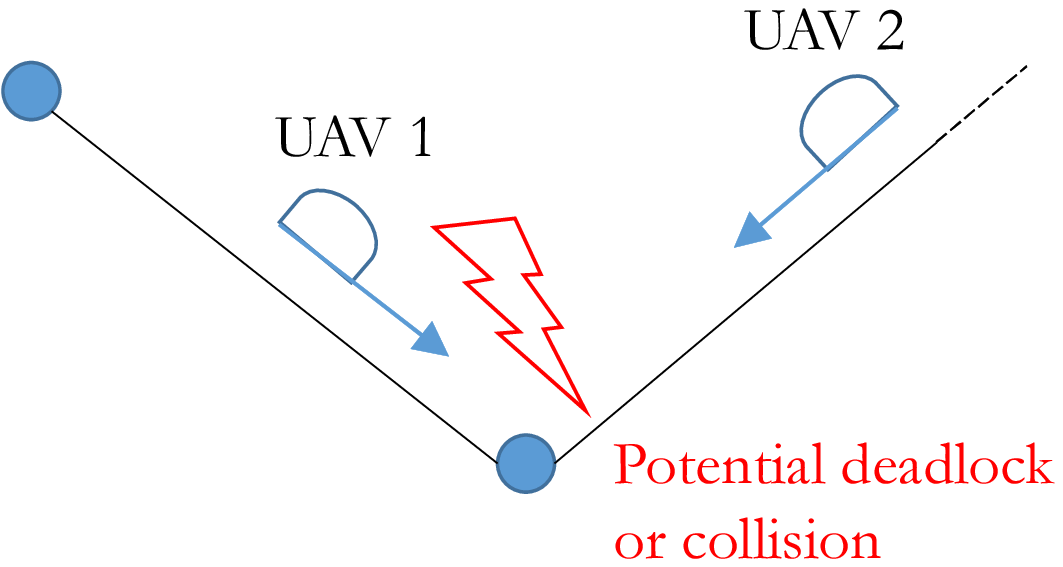}
  \caption{} \label{fig:nondir_path}
  \end{subfigure}
  \begin{subfigure}{0.4\textwidth}
  \includegraphics[width=\linewidth]{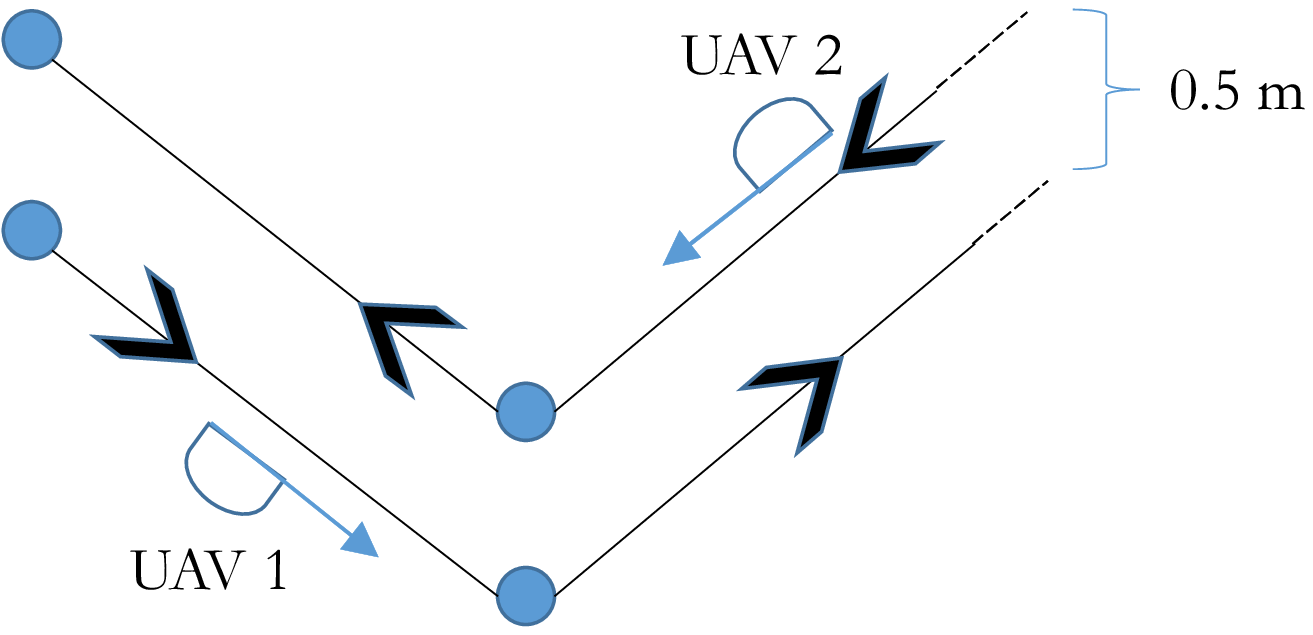}
  \caption{} \label{fig:dir_path}
  \end{subfigure}
  \caption{Potentially problematic non-directed paths environment (\ref{fig:nondir_path}) and a more robust directed paths environment (\ref{fig:dir_path})}\label{fig:paths}
\end{figure}

The scenarios of the UAV operations are conducted in both lab scale and industrial scale indoor environments. The lab scale indoor environment is depicted in Figure \ref{fig:env}. Each location of interest is indicated by a letter, where positions in the same location are distinguished by a number (following the letter). The green nodes are the recharge locations, where recharge actions take place. The established directed paths (illustrated in Figure \ref{fig:dir_path}) are also designed to avoid potential deadlock or collision. This means, with a sensor detecting the area forward where the UAV is going to fly, a short wait upon another UAV ahead will avoid collision among UAVs. In another word, such a situation in Figure \ref{fig:nondir_path} does not happen, where the UAVs collide or wait for one another continuously (deadlock).

The constraints of the task scheduling problem in this study are mentioned as follows.
\begin{enumerate}[topsep=0pt]
\item A task is executed once by one UAV.
\item A task is executed within the given time window.
\item A recharge station can be occupied by multiple UAVs.\\
The recharge station has no limited space due to the technology advancement in the space efficient battery recharge or even replacement platform \cite{suzuki2012automatic}, and a small size of UAV fleet (e.g. $\leq$10) may cover the most instances of UAV application in indoor environment.
The recharge time for a depleted battery with the capacity of 20 minutes is 45 minutes.
\item A non-recharge-station position can be occupied by at most one UAV at a particular time.
\item A UAV can only execute at most one task at a time.
\item A UAV may only land at a recharge station.
\item The battery level of a UAV must never be depleted at any non-recharge-station position.
\item The battery of each UAV is fully charged in the beginning of the planning horizon.
\item A task can only be executed after all its predecessors are completed.
\item There is no cyclic nor redundant precedence relationship.
\end{enumerate}


\begin{flushleft}
\textit{Notations}\\
\end{flushleft}
$i,j$ : index of task\\
$k$ : index of UAV\\
$l, l'$ : index of position\\
$r$ : number of recharge stations\\
$h_i$ : the index of nearest recharge station from the end position of task $i$\\
$R$ : set of recharge stations; $R={1,2,..r}$\\
$N$ : set of tasks; $N={1,2,..n}$\\
$V$ : set of UAVs; $V={1,2,..v}$\\
$a\prime_i$ : release time of task $i$\\
$z\prime_i$ : due time of task $i$\\
$w_i$ : processing time of task $i$\\
$s_i$ : start position of task $i$\\
$s_0^k$ : initial position of UAV $k$\\
$e_i$ : end position of task $i$\\
$d_{ij}$ : task $i$ is preceded by task $j$\\
$t_{l}^{l'}$ : travel time from $l$ to $l'$\\
$c_l^{h_i}$ : travel time from $l$ to recharge station $h_i$\\\\
$\alpha$ : maximum battery capacity ($\alpha=1200$)\\
$\gamma$ : one full cycle of recharge time\\\\
\textbf{Decision variables}\\
$a_i$ : start time of task $i$\\
$z_i$ : end time of task $i$\\
$b_k^i$ : battery level of UAV $k$ before executing task $i$\\
\[
x_i^k =
  \begin{cases}
    1  & \quad \text{if task } i \text{ is executed by UAV } k\\
    0  & \quad \text{otherwise}
  \end{cases}
\]
\[
y_i^k =
  \begin{cases}
    1  & \quad \text{if recharge is done by UAV } k \text{ before executing task }i\\
    0  & \quad \text{otherwise}
  \end{cases}
\]
\[
o_{ij} =
  \begin{cases}
    1  & \quad \text{if task } i \text{ is operationally preceded by task } j\\
    0  & \quad \text{otherwise}
  \end{cases}
\]
\[
p_i =
  \begin{cases}
    1  & \quad \text{if task } i \text{ is operationally preceded by another task}\\
    0  & \quad \text{otherwise}
  \end{cases}
\]
\[
x\prime_{ij}^k =
  \begin{cases}
    1  & \quad \text{if task } i \text{ and } j \text{ are executed by UAV } k\\
    0  & \quad \text{otherwise}
  \end{cases}
\]
\[
q_{ij} =
  \begin{cases}
    1  & \quad \text{if task } i \text{ and } j \text{ are executed by the same UAV}\\
    0  & \quad \text{otherwise}
  \end{cases}
\]
\[
u_{ij}^k =
  \begin{cases}
    1  & \quad \parbox[t]{.7\textwidth}{\raggedright
    if a recharge is conducted by UAV $k$ before task $i$ is executed and task $i$ is operationally preceded by task $j$}\\
    0  & \quad \text{otherwise}
  \end{cases}
\]
\[
v_{ij}^k =
  \begin{cases}
    1  & \quad \parbox[t]{.7\textwidth}{\raggedright
    if no recharge is conducted before task $i$ is executed by UAV $k$ and task $i$ is operationally preceded by task $j$}\\
    0  & \quad \text{otherwise}
  \end{cases}
\]
\[
m_{i}^k =
  \begin{cases}
    1  & \quad \text{if a flight is required from } s_0^k \text{ before UAV } k \text{ executes task } i\\
    0  & \quad \text{otherwise}
  \end{cases}
\]\\
$f_j$ : start time of a task which is operationally preceded by task $j$\\
$f\prime_{ij}^k$ : $a_i$ which is penalized with a large value $M$ when $a_i \le z_j$ or task $i$ (or $j$) is not executed by UAV $k$\\
\[
f\prime\prime_{ij} =
  \begin{cases}
    1  & \quad \text{if } a_i < z_j\\
    0  & \quad \text{otherwise}
  \end{cases}
\]\\
$g_{ij}^k$ : battery consumption when no recharge takes place before the execution of task $i$, which is preceded by $j$, by UAV $k$\\
$E$ : total battery consumption\\
Note: Task $i$ is operationally preceded by task $j$ if task $i$ is scheduled to be executed right after task $j$ by the same UAV; it is different from precedence rule.

\newpage

Objective function: minimize $E$\\

Subject to:\\
1 - end time is the summation of start time and processing time of the task\\
\begin{equation}\label{eq:constraint1}
  z_i = a_i + w_i \quad \forall i \in N
\end{equation}
2 - the start time and end time of a task shall not violate its time window\\
\begin{equation}\label{eq:constraint36}
  a_i \geq a\prime_i \quad \forall i \in N
\end{equation}
\begin{equation}\label{eq:constraint37}
  z_i \leq z\prime_i \quad \forall i \in N
\end{equation}
3 - every task is executed once by one UAV\\
\begin{equation}\label{eq:constraint2}
  \sum_{k \in V}x_i^k = 1 \quad \forall i \in N
\end{equation}
4 - a UAV can only execute one task at a time
\begin{equation}\label{eq:constraint3}
a_i \geq z_j \| z_i \leq a_j \quad \forall {i,j} \in N, i \neq j, x\prime_{ij}^k=1
\end{equation}
\begin{equation}\label{eq:constraint4}
x\prime_{ij}^k \geq x_i^k + x_j^k - 1 \quad \forall {i,j} \in N, i \neq j, \forall k \in V
\end{equation}
\begin{equation}\label{eq:constraint5}
x\prime_{ij}^k \leq x_i^k \quad \forall {i,j} \in N, i \neq j, \forall k \in V
\end{equation}
\begin{equation}\label{eq:constraint6}
x\prime_{ij}^k \leq x_j^k \quad \forall {i,j} \in N, i \neq j, \forall k \in V
\end{equation}
\begin{equation}\label{eq:constraint7}
q_{ij} = \sum_{k \in V}x\prime_{ij}^k \quad \forall {i,j} \in N
\end{equation}
\begin{equation}\label{eq:constraint8}
y_i^k \leq x_i^k \quad \forall i \in N, \forall k \in V
\end{equation}
5 - battery condition that will trigger recharge as necessary\\
\begin{equation}\label{eq:constraint9}
  b_k^i - (w_i + c_{e_i}^{h_i}) x_i^k \geq 0 \quad \forall i \in N, h_i \in R
\end{equation}
\begin{equation}\label{eq:constraint14}
  b_k^j - (z_i-a_j) - c_{e_i}^{h_i} \geq 0 \quad \forall {i,j} \in N, \forall k \in V, v_{i,j}^k=1
\end{equation}
6 - no multiple tasks are executed at the same position at a particular time\\
\begin{equation}\label{eq:constraint10}
  a_i \geq z_j || z_i \leq a_j \quad \forall {i,j} \in N, \lfloor\min(s_i, s_j)/\max(s_i,s_j)\rfloor = 1, i \neq j
\end{equation}
7 - define what recharge does to the battery level and how the battery is consumed\\
\begin{equation}\label{eq:constraint15}
  b_k^i = (\alpha-c_{s_i}^{h_j}) \quad \forall {i,j} \in N, u_{ij}^k=1\\, \forall j \in N, h_j \in R
\end{equation}
\begin{equation}\label{eq:constraint16}
  b_k^i = b_k^j-(a_i-a_j) \quad \forall {i,j} \in N, v_{ij}^k=1
\end{equation}
\begin{equation}\label{eq:constraint11}
u_{ij}^k \geq o_{ij} + y_i^k - 1 \quad \forall {i,j} \in N, \forall k \in V
\end{equation}
\begin{equation}\label{eq:constraint12}
u_{ij}^k + v_{ij}^k \leq o_{ij} \quad \forall {i,j} \in N, \forall k \in V
\end{equation}
\begin{equation}\label{eq:constraint13}
v_{ij}^k \geq o_{ij} + x_i^k - y_i^k - 1 \quad \forall {i,j} \in N, \forall k \in V
\end{equation}
\begin{equation}\label{eq:constraint34}
g_{ij}^k \geq z_i-a_j \quad \forall {i,j} \in N, \forall k \in V, v_{ij}^k=1
\end{equation}
8 - set the battery to be consumed for the flight from the UAV's initial position towards the first task; which is an exception case to the defined constraints for battery consumption in \ref{eq:constraint15} and \ref{eq:constraint16}.
\begin{equation}\label{eq:constraint17}
b_k^i = \alpha - t_{s_0^k}^{s_i} \quad \forall {i,j} \in N, \forall k \in V, p_i=0, x_i^k=1
\end{equation}
\begin{equation}\label{eq:constraint35}
m_i^k \geq x_i^k - p_i \quad \forall {i,j} \in N, \forall k \in V, p_i=0, x_i^k=1
\end{equation}
9 - determine the start time of a task which is the first task of a particular UAV
\begin{equation}\label{eq:constraint18}
t_{s_0^k}^{s_i}*m_i^k \leq a_i \quad \forall i \in N, \forall k \in V
\end{equation}
10 - operational precedence relationship indicates task $i$ is executed right after task $j$ by the same UAV; $j$ operationally precedes $i$
\begin{multline}\label{eq:constraint19}
o_{ij} \geq -a_i + f_j - f\prime\prime_{ij} 2M - (2-x_i^k-x_j^k) 2M + 1\\
\forall {i,j} \in N, \forall k \in V
\end{multline}
\begin{equation}\label{eq:constraint20}
o_{ij} \leq q_{ij} \quad \forall {i,j} \in N
\end{equation}
\begin{equation}\label{eq:constraint21}
f_j = \min\limits_{i \in N, k \in V} f\prime_{ij}^k \quad \forall j \in N
\end{equation}
\begin{equation}\label{eq:constraint22}
f\prime_{ij}^k = a_i + f\prime\prime_{ij} M + (2-x_i^k-x_j^k) M \quad \forall {i,j} \in N, \forall k \in V
\end{equation}
\begin{equation}\label{eq:constraint23}
f\prime\prime_{ij} = 1-(a_i \geq z_j) \quad \forall {i,j} \in N
\end{equation}
11 - a task can operationally precede and be operationally preceded by at most one other task
\begin{equation}\label{eq:constraint25}
\sum_{i \in N}(1-p_i) \leq v
\end{equation}
\begin{equation}\label{eq:constraint28}
\sum_{i \in N}o_{ij} \leq 1 \quad \forall j \in N
\end{equation}
\begin{equation}\label{eq:constraint24}
p_i = \sum_{j \in N}o_{ij} \quad \forall i \in N
\end{equation}
12 - no self operational-precedence
\begin{equation}\label{eq:constraint26}
o_{ii}=0 \quad \forall i \in N
\end{equation}
13 - no cyclic operational-precedence
\begin{equation}\label{eq:constraint27}
o_{ij} + o_{ji} \leq 1 \quad \forall {i,j} \in N
\end{equation}
14 - guaranteed feasible recharge time in the schedule\\
\begin{equation}\label{eq:constraint29}
  z_j + t_{e_j}^{s_i}(v_{ij}^k) + (c_{e_j}^{h_j}+\gamma+c_{s_i}^{h_j})u_{ij}^k \leq a_i \quad \forall {i,j} \in N, \forall k \in V, o_{ij}=1, \forall h_j \in R
\end{equation}
15 - precedence constraint\\
\begin{equation}\label{eq:constraint30}
  a_i \geq z_j d_{ij} \quad \forall {i,j} \in N
\end{equation}
16 - the sum of all UAVs' battery consumption does not exceed the total battery consumption
\begin{equation}\label{eq:constraint31}
\sum_{i \in N}\sum_{k \in V}{((\sum_{j \in N}{g_{ij}^k + u_{ij}^k*(c_{e_j}^{h_j}+w_i+c_{s_i}^{h_j})}) + m_i^k*(t_{s_0^k}^{s_i}+w_i))} \leq E
\end{equation}

The developed MILP model is solved for small-scale problems, and the data is shown later in Section \ref{sec:sec_num_exp}.
The exponentially growing computation time indicates the need of a heuristic-based approach which is capable of solving the large-scale problem within a reasonable amount of time.
Such a methodology is developed in this study, and it is discussed in the following section.

\newpage
\section{Methodology}
\label{sec:sec_methodology}
The proposed Restful Task Assignment Algorithm is described through Algorithm \ref{algo:algo_prefUAVs}\mytilde \ref{algo:algo_fragment_asap} later in this section. To help to minimize the battery consumption while parallelizing task executions when deemed fit (which may lead to a shorter total execution time), Algorithm \ref{algo:algo_prefUAVs} is presented.
The time window attribute on the task encourages the usage of Algorithm \ref{algo:algo_fragment_alap}: Backward Fragment Placement Algorithm (BFPA) to quickly find a feasible solution if it exists and increase the system's robustness. Moreover, this method tends to preserve the battery level, since some empty time ranges in between the task executions can be used for having recharge actions. The detailed procedure is described in Section \ref{sec:subsec_rtaa}. A support for assigning tasks which couldn't be assigned in BFPA and those without time window is provided by Algorithm \ref{algo:algo_fragment_asap}: Forward Fragment Placement Algorithm (FFPA). In respect to the route query between positions, Dijkstra's Algorithm \cite{dijkstra1959note} is employed in the map module.

RTAA is then incorporated with Particle Swarm Optimization (PSO), where task sequences from the initial swarm and the produced mutations throughout the whole search iterations are the inputs for RTAA. The mechanism of PSO is depicted in Section \ref{sec:subsec_pso}, followed by the description of RTAA in detail in Section \ref{sec:subsec_rtaa}.

\subsection{Particle Swarm Optimization Algorithm}
\label{sec:subsec_pso}
PSO is an optimization algorithm which falls under the category of metaheuristic approach. It allows explorations to areas which may not look promising in the beginning of the search, alienating it from getting trapped in a local optimum solution. The incorporation of PSO performs the battery consumption minimization process during a reasonable computation time. The calculation of the total battery consumption of a schedule is contributed only by task executions, flight actions (among positions of task execution and recharge stations), and hover actions. The PSO algorithm is depicted in Algorithm \ref{algo:algo_pso}, where more details of the incorporation with such a problem nature of this paper can be found in \cite{khosiawan2016task}.

\begin{algorithm} [htp]
\caption{Particle Swarm Optimization Algorithm}\label{algo:algo_pso}
\begin{algorithmic}[1]
\Statex \textbf{Input:} Initial Swarm ($swarm$)
\Statex \textbf{Output:} schedule of tasks on UAVs ($schedule$)
\State Initialize (parameters, swarm, local best and global best)
\While {stop condition not met}
\State $velocity \leftarrow$  updateVelocity(swarm, velocity, local best, global best);
\State $swarm \leftarrow$ updateSwarm(swarm, velocity);
\State $localBest \leftarrow$ getLocalBest(fitness(swarm), localbest);
\State $globalBest \leftarrow$ getGlobalBest(localBest, globalBest);
\State generation++;
\EndWhile
\end{algorithmic}
\end{algorithm}

The initial swarm has a role in placing initial starting points \cite{book:958913} throughout the solution space. There are 10 priority rules used in the initial swarm generation. Eight of them are addressed in \cite{khosiawan2016task}, while two other rules in regard to the position occupation are introduced in this paper. Those are most and less occupied position rules, task will be sorted based on the projected-occupation-load of its execution position in ascending and descending manner. For material handling task, these two rules see the start position as the execution position. The example of sequences (particles) in the initial swarm created through the priority rules in regard to the dataset in Table \ref{table:tasks} is depicted in Table \ref{table:initial_swarm}.

\begin{table} [ht]
\caption{Priority rules for initial particle generation}
\begin{center}
    \begin{tabularx}{\textwidth}{X|c|c|c|c|c|c|c|c|c|c}
    \hline\noalign{\smallskip}
    Heuristic Rules & \multicolumn{10}{c}{Task Sequence (Task ID)}\\
    \noalign{\smallskip}\hline\noalign{\smallskip}
    Minimum Number of Cumulative Predecessors & 5 & 8 & 1 & 9 & 10 & 2 & 4 & 7 & 3 & 6 \\ \hline
    Minimum Total Number Of Predecessors & 5 & 8 & 1 & 2 & 3 & 4 & 6 & 9 & 10 & 7 \\ \hline
    Maximum Number of Cumulative Successors & 8 & 5 & 9 & 10 & 7 & 1 & 2 & 3 & 4 & 6 \\ \hline
    Maximum Total Number of Successors & 7 & 8 & 9 & 10 & 5 & 1 & 2 & 3 & 4 & 6 \\ \hline
    Maximum Task Execution Time & 10 & 4 & 3 & 6 & 9 & 1 & 2 & 5 & 7 & 8 \\ \hline
    Minimum Task Execution Time & 1 & 2 & 5 & 7 & 8 & 6 & 9 & 3 & 4 & 10 \\ \hline
    Maximum Ranked Positional Weight & 5 & 8 & 9 & 10 & 7 & 1 & 2 & 3 & 4 & 6 \\ \hline
    Minimum Inverse Positional Weight & 5 & 8 & 1 & 9 & 10 & 4 & 2 & 7 & 3 & 6 \\ \hline
    Tasks with Less Occupied Position & 1 & 8 & 7 & 2 & 9 & 6 & 4 & 10 & 3 & 5 \\ \hline
    Tasks with Most Occupied Position & 3 & 5 & 10 & 4 & 6 & 9 & 2 & 7 & 8 & 1 \\ \hline
    \end{tabularx}
\end{center}
\label{table:initial_swarm}
\end{table}

%
%
%
%
\subsection{Restful Task Assignment Algorithm}
\label{sec:subsec_rtaa}
The scheduling algorithm is depicted by Algorithm \ref{algo:algo_fragment}\mytilde\ref{algo:algo_fragment_asap}. The encapsulation of sub-methods performed in Algorithm \ref{algo:algo_prefUAVs}\mytilde\ref{algo:algo_fragment_asap} provides a more clear top-constitutional-view of the heuristic approach depicted in Algorithm \ref{algo:algo_fragment}: Restful Task Assignment Algorithm (RTAA). The main idea of RTAA is to promote the existence of a continuous free timespan which can be used by the UAV to have a recharge.
The algorithm takes input from a sequence of tasks created initially through the priority rules or update of position during a particular iteration. In line 1, the tasks are grouped based on the position occupation time where the task will start. For inspection tasks, it also acts as the position where the task will take place. Through the formed group, the positions will be ranked in a descending order; from the most occupied one to the most idle one. From line 3\mytilde22, a method of creating a preliminary schedule without recharge action is depicted.
Based on the ranked position list, tasks in the group will be iterated and tried to be assigned to a UAV to be executed at a particular time range. This provides the structure of the schedule by putting the tasks which are less flexible (i.e. to be executed at one position as many others may yield a smaller execution time range). In line 7, the task is put into schedule in a backward manner. Following the successful assignment, the record of UAV occupation-timestamp fragments (UOF) and position occupation-timestamp fragments (POF) are updated.

\begin{figure}[H]
  \centering
  \includegraphics[width=0.8\linewidth]{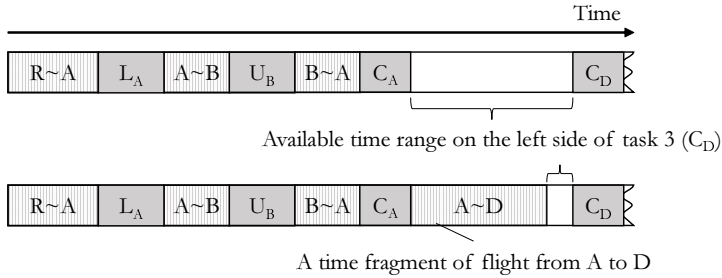}
  \caption{A time fragment put into the available time range based on the calculation according to a part of the scheduling algorithm}
  \label{fig:time}
\end{figure}

Tasks which couldn't be assigned previously and those without time windows will be attempted to be scheduled in line 14\mytilde22. Here, the tasks are scheduled on the available time range according to the UOF and POF. When there are multiple possible placements of time fragment in the available time range, the task is assigned at the soonest execution time. If there is a task which still can't be assigned in this attempt, a return value is returned indicating that another sequence should be checked since the current one does not yield a complete feasible schedule for the given set of tasks. In this manner, assuming that the given tasks are rational, the computation time can be minimized. The distinction of time fragment and time range is depicted in Figure \ref{fig:time}.

After the preliminary schedule is constructed, the insertion of recharge, hover, and wait-on-ground actions is performed (line 23\mytilde 30). In this regard, a minimum recharge time shall be defined, to determine the amount of time that is worth (long enough) to be used for a recharge action. On the other hand, if the minimum recharge time is too long, then the potential time that actually can be used for recharging the battery would be wasted. The recharge insertion procedure looks for available time ranges which are not used for any task execution and at least equal to or greater than the defined minimum recharge time.

In every attempt to put a task into the schedule, indicated by '+' throughout the pseudocode (Algorithm \ref{algo:algo_fragment_alap} line 8$^+$ and Algorithm \ref{algo:algo_fragment_asap} line 7$^+$), there is a check of the required flight's execution time in respect to the adjacent task(s)'s and possibility to put a recharge action when applicable or necessary. This means, there will be no insufficient time for the UAV to fly from the end position of a preceding task to the start position of the successor task. Furthermore, there will be no insufficient battery value since a task will not be assigned at a particular time fragment when the formed sequence of tasks does not leave enough battery to go at least to the nearest recharge station. During this process, the time fragment for the task execution is ensured not to overlap with the existing UAV occupation time fragments. The reason why it is not done together with the trimming based on the position occupation time fragments is because of the dynamic flights of the UAVs between positions (where a recharge action might be replaced by a hover and a task execution \textemdash at a latter task assignment) causing frequent changes throughout the overall schedule construction. On the other hand, position occupation time fragments are stagnant once the assignments are made, regardless of the previous or the following assignments.

\begin{figure}[H]
  \centering
  \begin{subfigure}{0.5\textwidth}
  \includegraphics[width=\linewidth]{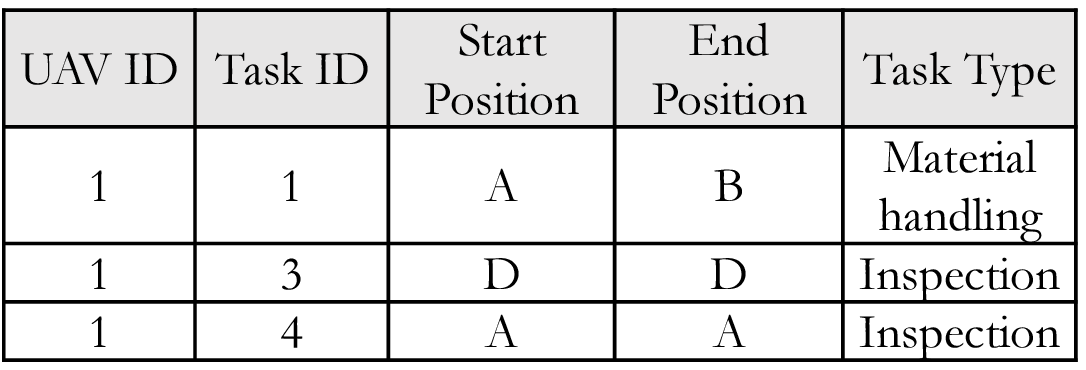}
  \caption{Information on a sample set of tasks} \label{fig:tasks}
  \end{subfigure}
  \begin{subfigure}{0.8\textwidth}
  \includegraphics[width=\linewidth]{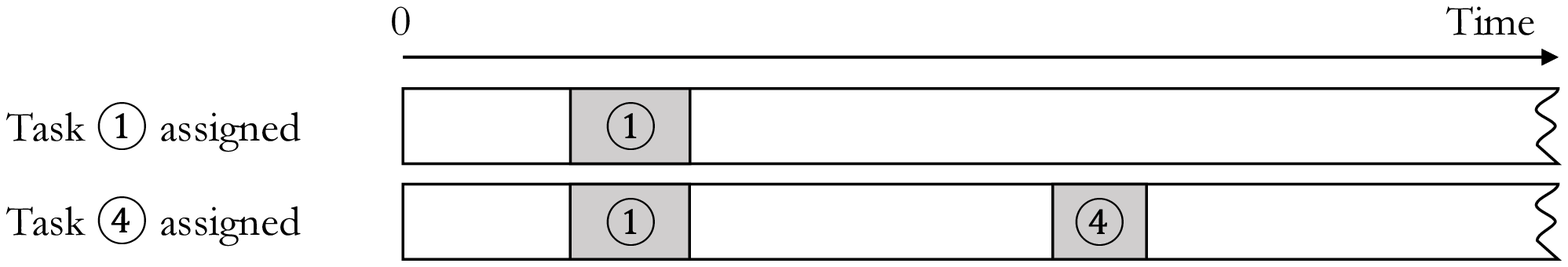}
  \caption{Occupation time fragments of position A} \label{fig:pos_occ}
  \end{subfigure}
  \begin{subfigure}{0.8\textwidth}
  \includegraphics[width=\linewidth]{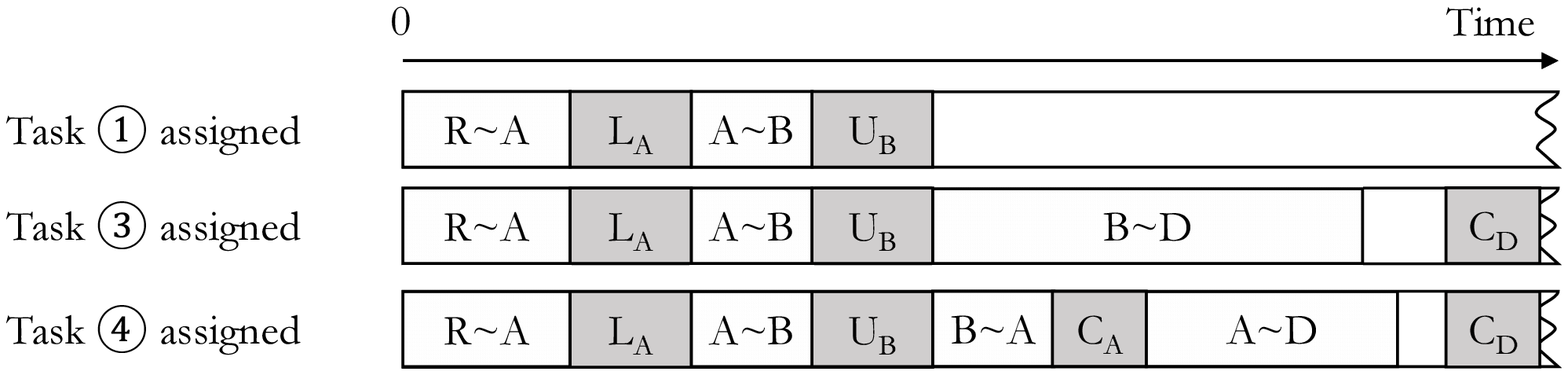}
  \caption{Occupation time fragments of UAV 1} \label{fig:uav_occ}
  \end{subfigure}
  \caption{Position and UAV occupation time fragments}\label{fig:pos_uav_occ}
\end{figure}

The distinction on the characteristic of UAV occupation and position occupation fragments is depicted in Figure \ref{fig:pos_uav_occ}. Let's assume that during the schedule construction, several task assignments are performed, where task \circled{1}, \circled{3}, and \circled{4} are assigned sequentially. In figure \ref{fig:pos_occ}, a particular position occupation time fragment is depicted to be stagnant once the respective task is assigned. However, the UAV occupation task is depicted otherwise in Figure \ref{fig:uav_occ}. In the beginning, a material handling task \circled{1} is assigned\textemdash where the time fragments for flight from the recharge station to $A$ ($R$\mytilde $A$), a load ($L_A$), flight from $A$ to $B$ ($A$\mytilde $B$), and unload ($U_B$) are put at their calculated time ranges. When task \circled{3} is assigned to the schedule, assuming that the related time-fragments satisfy the considered constraints, a time fragment of flight from the end position of task \circled{1}: $B$ to the start position of task \circled{3}: $D$ is assigned to the respective time-range. Afterwards, task \circled{4} is assigned according to its time window, and it occupies the time range between the end of task \circled{1} and the start of task \circled{3}. This means that the time fragment of flight B\mytilde D is obsolete now. Instead, fragments of flight $B$\mytilde $A$, capturing inspection image at $A$, and flight $A$\mytilde $D$ are assigned in the respective time range. In another word, every assigned occupation time fragment of UAV 1 may change over time during the schedule construction.

\begin{algorithm}[H]
\caption{Restful Task Assignment Algorithm}\label{algo:algo_fragment}
\begin{algorithmic}[1]
\Statex \textbf{Input:} sequence of tasks ($seq$)
\Statex \textbf{Output:} schedule of task executions ($sched$)

\State $taskGroup \leftarrow$ groupTasksBasedOnStartPositions($seq$)
\State $posRank \leftarrow$ rankPositionUsageFrequencyFromTheBusiestOne($taskGroup$)
\Statex
\Statex \textbf{Create schedule without recharge action:}
\For {$position$ in $posRank.sortedList$}
\State $tasks \leftarrow taskGroup$.get($position$)
\For {$t$ in $tasks$}
\If {hasTimeWindow($t$)}
\State $isScheduled \leftarrow$ BFPA($posOccupationFragments$, $uavOccupationFragments$, $sched$, $t$)
\If {$isScheduled$}
\State $seq$.remove($t$)
\EndIf
\EndIf
\EndFor
\EndFor

\For {$t$ in $seq$}
\State $isScheduled \leftarrow$ FFPA($posOccupationFragments$, $uavOccupationFragments$, $sched$, $t$)
\If {$isScheduled$}
\State return getTotalBatteryConsumption($schedule$)
\Else
\State NO\_SCHEDULE $\leftarrow$ -1
\State return NO\_SCHEDULE
\EndIf
\EndFor

\Statex
\Statex \textbf{Insert recharge and other actions:}
\State $utfs \leftarrow$ getUnallocatedTimeFragments($sched$)
\For {$utf$ in $utfs$}
\If {$utf \leq$  MIN\_RECHARGE\_TIME}
\State putActionsForRecharge($sched$, $utf$)
\Else
\State putActionsForFlightBeetwenPositions($sched$, $utf$)
\EndIf
\EndFor
\end{algorithmic}
\end{algorithm}

Prior to the task assignment, a sorting of the preferred UAV list for performing a particular task is done. This process is done based on the precedence relationships of the task to be assigned (1,2,4,5) and the possessed workload (line 3). The motivation behind the utilization of precedence relationship here is the tendency of the successor tasks being able to be executed immediately after its latest predecessor is done.

\begin{algorithm}[H]
\caption{Sort the preferred UAV list}\label{algo:algo_prefUAVs}
\begin{algorithmic}[1]
\Statex \textbf{Input:} a list of all UAV IDs ($uavIDs$), UAV occupation fragment ($uavOccFragments$), current schedule ($sched$), upcoming task to be assigned ($t$)
\Statex \textbf{Output:} a sorted list of UAV IDs \textemdash from the most preferred one to the least ($prefUAVs$)

\State $uavOnDutyBefore \leftarrow$ getUAVofTheLatestPredecessorIfExist($t$, $schedule$)
\State $uavOnDutyAfter \leftarrow$ getUAVofTheEarliestSuccessorIfExist($t$, $schedule$)

\State $prefUAVs \leftarrow$ sortUAVIDsBasedOnCurrentWorkload($uavOccFragment$, $uavIDs$)

\State moveUAVIDToTheFrontOfList($prefUAVs$, $uavOnDutyAfter$)
\State moveUAVIDToTheFrontOfList($prefUAVs$, $uavOnDutyBefore$)

\State return $prefUAVs$
\end{algorithmic}
\end{algorithm}

The details of the first task assignment attempt is depicted in Algorithm \ref{algo:algo_fragment_alap}: Backward Fragment Placement Algorithm (BFAP). In line 4\mytilde7, possible time fragments for the task to be assigned at are listed. They are obtained by trimming the fragments of the original time window of the given task which overlap with the current record of POF. The trimming based on the end time of a predecessor and the start time of a successor is a consequence of the design of Algorithm \ref{algo:algo_fragment}, where a left to right (time-wise) task assignment procedure is not a must.

\begin{figure}[H]
  \centering
  \begin{subfigure}{0.5\textwidth}
  \includegraphics[width=\linewidth]{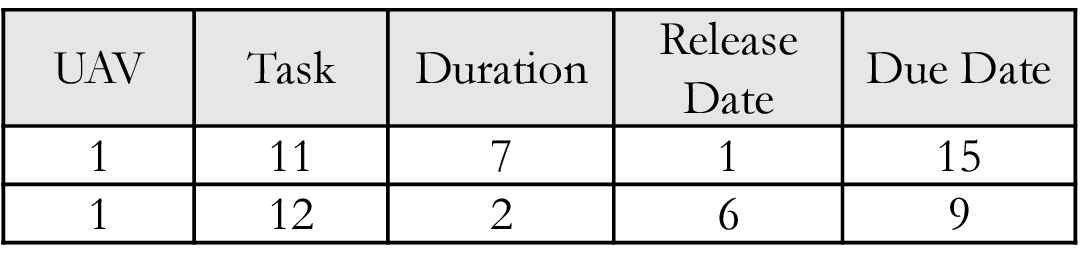}
  \caption{Tasks with time window; \circled{12} has less slack than \circled{11}} \label{fig:alap_tasks}
  \end{subfigure}
  \begin{subfigure}{0.8\textwidth}
  \includegraphics[width=\linewidth]{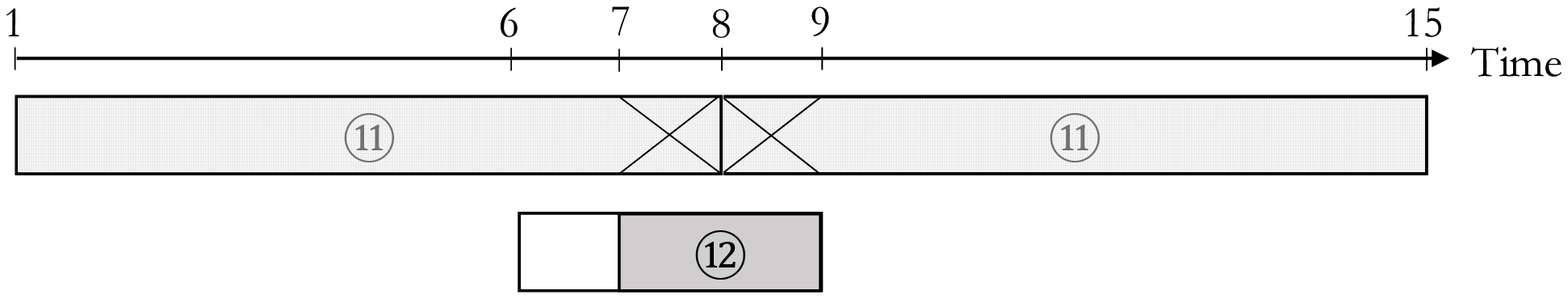}
  \caption{Formerly assign (based on the manner described in Algorithm \ref{algo:algo_fragment_alap}) task \circled{12} and then \circled{11} afterwards} \label{fig:alap_not_ok}
  \end{subfigure}
  \begin{subfigure}{0.8\textwidth}
  \includegraphics[width=\linewidth]{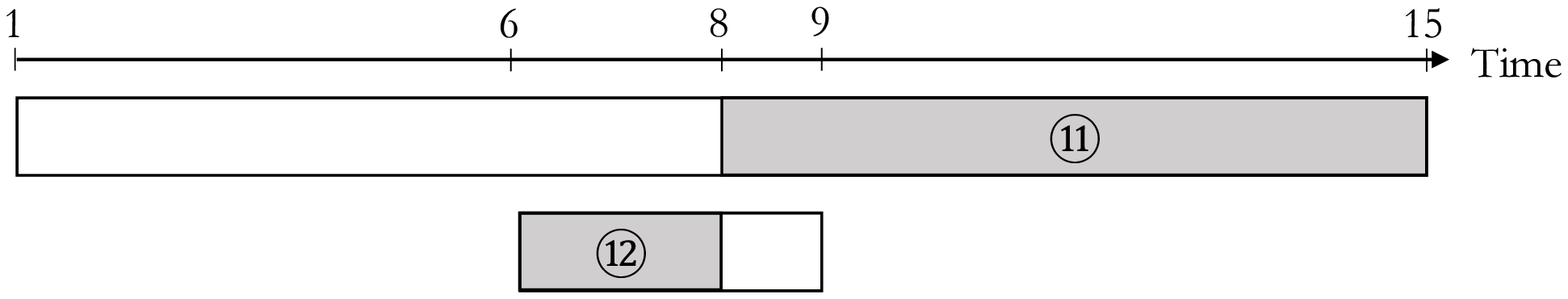}
  \caption{Formerly assign (based on the manner described in Algorithm \ref{algo:algo_fragment_alap}) task \circled{11} and then \circled{12} afterwards} \label{fig:alap_ok}
  \end{subfigure}
  \caption{A task with more slack may actually be formerly assigned to yield a feasible solution in BFAP }\label{fig:alap_variation}
\end{figure}

In line 8$^+$, the task is attempted to be assigned to the schedule in a backward manner to give an enhanced number of possibilities for the other tasks to be arranged without violating the given time window constraints. Once the task is assigned, the occupation fragments of the respective UAV and position are updated (line 10).
Furthermore, when assigning tasks with the same precedence level, one with less slack may need to be assigned before the other(s) or maybe the other way around to yield a feasible schedule - as depicted in Figure \ref{fig:alap_variation}. Task \circled{12} has less slack than \circled{11} and one may prefer to assign \circled{12} before \circled{11}. This results in the next task assignment attempt becoming infeasible (depicted in Figure \ref{fig:alap_not_ok}), while the other way around in Figure \ref{fig:alap_ok} is feasible. This is where the role of task sequence representation in the incorporated Particle Swarm Optimization comes in. The various task sequences enable a flexible exploration even towards an unlikely area. Additionally, it also promotes a quick exploration in a large solution space. In this manner, when there is no feasible schedule found, one can conclude that the given tasks (with the defined time windows) are too tight for the available resources (i.e., UAVs and positions). This is also supported by the condition where the tasks are not ranked based on their slack time before the task assignment - such treatment demotes the exploration during the search.

\begin{algorithm}[H]
\caption{Backward Fragment Placement Algorithm}\label{algo:algo_fragment_alap}
\begin{algorithmic}[1]
\Statex \textbf{Input:} position occupation time fragments ($posOccupationFragments$), UAV occupation time fragments ($uavOccupationFragments$), schedule of task executions ($sched$), task to be scheduled ($t$)
\Statex \textbf{Output:} scheduling completion status
\If {hasNoTimeWindow($t$)}
\State return false
\EndIf
\Statex
\Statex \textbf{List possible time fragments for the task to be assigned at based on the existing position occupation time fragments:}
\State $prefUAVs \leftarrow$ getPreferredUAVs($uavOccupationFragments$, $sched$, $t$) \Comment{Get a list of UAVs sorted based on defined preferences: executor of a predecessor of task $t$, executor of a follower of task $t$, least loaded UAV}
\State $predBasedTimeWin \leftarrow$ trimTimeWin($t$.timeWindow, $sched$) \Comment{Trim time window based on the scheduled predecessor(s) and follower(s)}
\State $freeStartPosFragments \leftarrow$ trimOccupiedTimeRange($predBasedTimeWin$, $posOccupationFragments$.get($t$.startPos))
\State $freeEndPosFragments \leftarrow$ trimOccupiedTimeRange($predBasedTimeWin$, $posOccupationFragments$.get($t$.endPos))
\Statex
\Statex \textbf{Attempt to put the task into the schedule in a backward manner:}
\State $isScheduled \leftarrow$ assignTaskBackward($freeStartPosFragments$, $freeEndPosFragments$, $prefUAVs$, $sched$, $t$)$^+$
\If{$isScheduled$}
\State updateOccupationFragments($posOccupationFragments$, $uavOccupationFragments$, $sched$) \Comment{Update position and UAV occupation fragments according to the recently scheduled task}
\EndIf
\Statex
\State return $isScheduled$
\end{algorithmic}
\end{algorithm}

After the attempt of assignment of tasks with time windows in Algorithm \ref{algo:algo_fragment_alap} is performed, the unassigned ones or tasks without time window will be attempted to be assigned based on Algorithm \ref{algo:algo_fragment_asap}: Forward Fragment Placement Algorithm. In this second attempt, several tasks with time window assigned in a tardy manner may be found when the respective time range allows the task to still meet the defined precedence relationships. However, on the opposite, a task with time window shall not be assigned before its release date. This assumption is made according to the accommodated task where the existence of lead task may fail the purpose of the task itself. For example, an inspection task is scheduled to be performed in the middle of the day: when a wind turbine blade would have been manufactured further. In such a situation, another (lead) inspection task executed in advance in the morning (before the given time window in the middle of the day) gives no contribution to the original intention of that task.

Generally, in the second attempt, the task assignment based on Algorithm \ref{algo:algo_fragment_asap} is attempted on the left side of a particular existing fragment (line 6). When there is a flexibility in the available time range, the soonest feasible time fragment will be chosen over the others (line 7). Such task assignment manner is promoted by the characteristic of Algorithm \ref{algo:algo_fragment_alap} where it tends to leave an empty area on the left-side of the (scheduled) task fragment. Hence, through Algorithm \ref{algo:algo_fragment_asap}, this condition will be evened out by the insertion of the remaining unassigned tasks and potential recharge actions.
In line 15, unscheduled tasks will be lastly attempted to be assigned after the last task in the existing schedule of a particular UAV when possible \textemdash satisfying the precedence relationships.

\begin{algorithm}[H]
\caption{Forward Fragment Placement Algorithm}\label{algo:algo_fragment_asap}
\begin{algorithmic}[1]
\State $prefUAVs \leftarrow$ getPreferredUAVs($uavOccupationFragments$, $sched$, $t$) \Comment{Get a list of UAVs sorted based on defined preferences: executor of a predecessor of task $t$, executor of a follower of task $t$, least loaded UAV}
\State $predBasedTimeWin \leftarrow$ trimTimeWin($t$.timeWindow, $sched$) \Comment{Trim time window based on the scheduled predecessor(s) and follower(s)}
\Statex
\Statex \textbf{Attempt to put the task into the schedule in a forward manner:}
\State TRY\_FFPA:
\For{$uav$ in $prefUAVs$}
\For{$st$ in $sched$.get($uav$)}
\State $freeTimeFragment \leftarrow$ getFreeFragmentOnTheLeftSideOfScheduledTask($st$, $posOccupationFragment$) \Comment Based on position occupation time fragments
\State $isScheduled \leftarrow$ assignTaskForward($freeTimeFragment$, $sched$, $t$)$^+$
\If {$isScheduled$}
\State updateOccupationFragments($posOccupationFragments$, $uavOccupationFragments$, $sched$) \Comment{Update position and UAV occupation fragments according to the recently scheduled task}
\State break TRY\_FFPA
\EndIf
\EndFor
\EndFor
\Statex
\Statex \textbf{Last attempt to schedule the given tasks if all previous ones (including BFPA) failed:}
\If {!$isScheduled$}
\State $isScheduled$ = assignAfterLastTaskForward($prefUAVs$, $sched$, $t$)
\EndIf
\Statex
\State return $isScheduled$
\end{algorithmic}
\end{algorithm}

\section{Numerical Experiment}
\label{sec:sec_num_exp}
The developed MILP in Section \ref{sec:sec_problem} is attempted to be solved in IBM ILOG CPLEX, with 3 UAVs and up to 9 tasks. It is run on an Intel Core i7 processor (2.9 GHz) with 32 GB of RAM. The computation time of CPLEX grows exponentially as shown in Table \ref{table:cplex_runs}. It reaches more than 2.5 hours after only 8 tasks, and this is significantly longer than the heuristic-based computation time for 100 tasks (presented later in this section). This condition essentially drives the need of pursuing a heuristic-based approach to obtain a good quality solution in a reasonable time.
\begin{table} [htp]
\caption{Computation time of small scale problems solved in CPLEX}
\begin{center}
    \begin{tabular}{ l c c }
    \hline\noalign{\smallskip}
    Number of tasks & Execution time (s) & Solution \\
    \noalign{\smallskip}\hline\noalign{\smallskip}
    4 & 1.73 & optimum \\
    5 & 6.12 & optimum \\
    6 & 25.82 & optimum \\
    7 & 182.86) & optimum \\
    8 & 10000 (stopped) & feasible \\
    9 & 10000 (stopped) & feasible \\ \hline
    \end{tabular}
\end{center}
\label{table:cplex_runs}
\end{table}

\begin{figure}[H]
  \centering
  \includegraphics[width=0.5\linewidth]{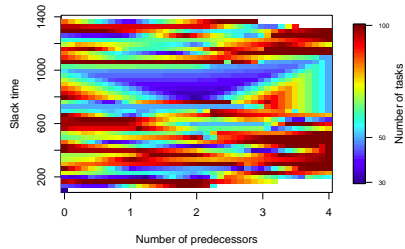}
  \caption{Summary of characteristics of the benchmark datasets}
  \label{fig:datasets_summary}
\end{figure}

The proposed methodology has been benchmarked by 54 datasets generated based on the flight demonstrations at the lab facility of the Department of Mechanical, Production and Management Engineering.
The datasets have different characteristics, formed by different levels of four parameters: geographical scale, number of tasks, predecessor distribution, and slack time distribution.
The summary of the task characteristics interpolated from both lab scale and industrial scale datasets can be illustrated in Figure \ref{fig:datasets_summary}.
\begin{enumerate}[topsep=0pt]
\item There are 2 geographical scales of the environment: lab scale and industrial scale with the ratio of 1:8.
The measurement of the lab environment is within 14 meters in width, 20 meters in length, and 7 meters in height.
\item For each geographical scale, there are 3 different numbers of tasks: 30, 50, and 100 tasks.
\item For each number of tasks, there are 3 predecessor distributions: 0, 1, 2.
As depicted in Figure \ref{fig:datasets_summary}, there exist tasks with 3 or 4 predecessors in a particular dataset.
The addressed parameter determines that the number of predecessors in a dataset satisfies the normal distribution with $\overline{x}=0\lor1\lor2$ and $\sigma=min(1, \overline{x})$ \textemdash not the exact number of predecessors for each task.
\item For each number of predecessor distribution, there are 3 slack time distributions: 300, 600, and 1200 seconds.
As depicted in Figure \ref{fig:datasets_summary}, there are some tasks with slack time of at least 100 and up to 1395 seconds.
The addressed parameter determines that the slack time in a dataset satisfies the normal distribution with $\overline{x}=300\lor600\lor1200$ and $\sigma=\overline{x}/5$ \textemdash not the exact slack time for each task.
\item Each dataset is used in a scheduling scenario which is run 20 times, whose results can provide a good quality observation.
\end{enumerate}

\begin{figure}[H]
  \centering
  \begin{subfigure}{0.3\textwidth}
  \includegraphics[width=\linewidth]{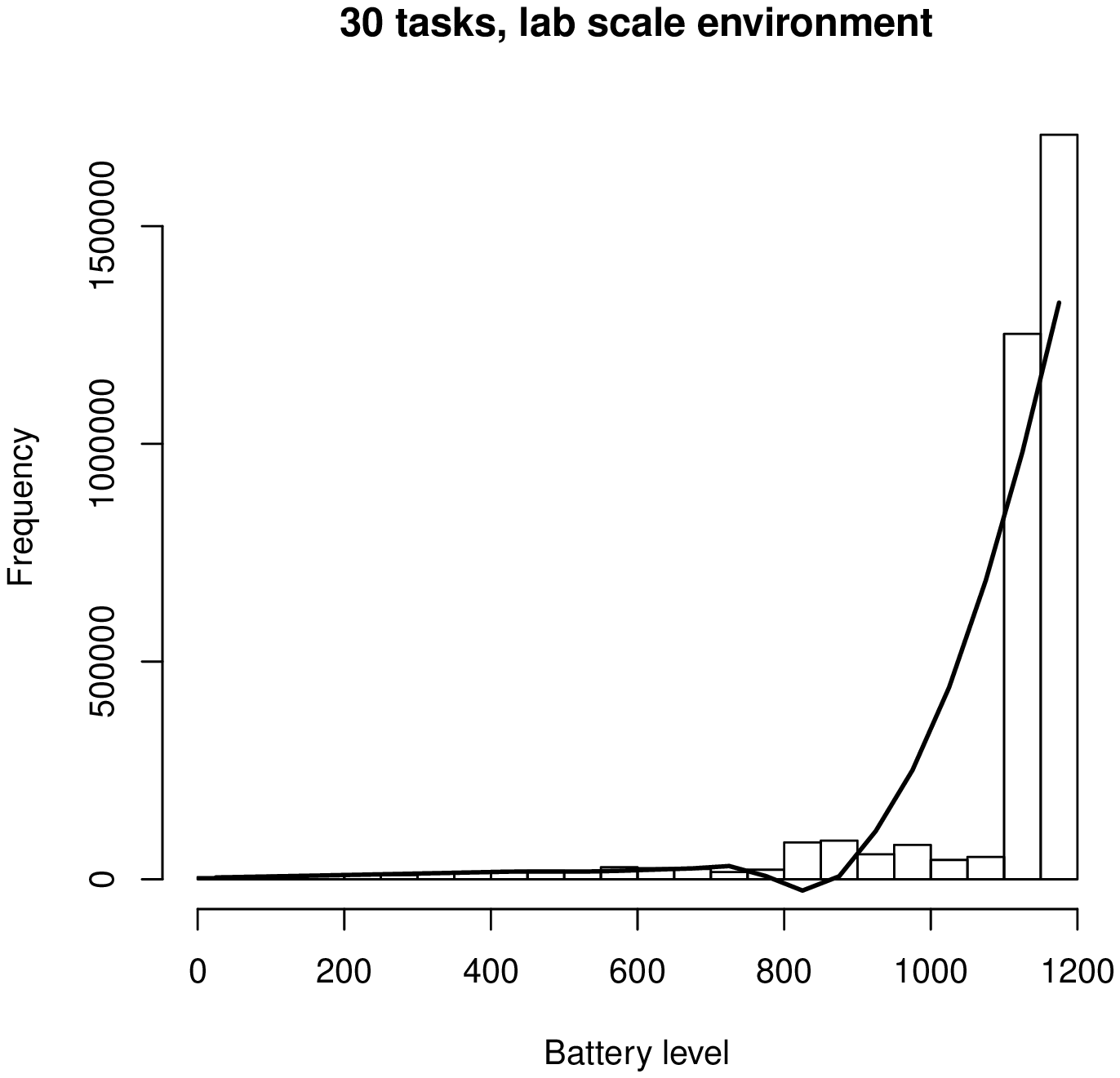}
  \caption{}\label{fig:battery_30_1}
  \end{subfigure}
  \begin{subfigure}{0.3\textwidth}
  \includegraphics[width=\linewidth]{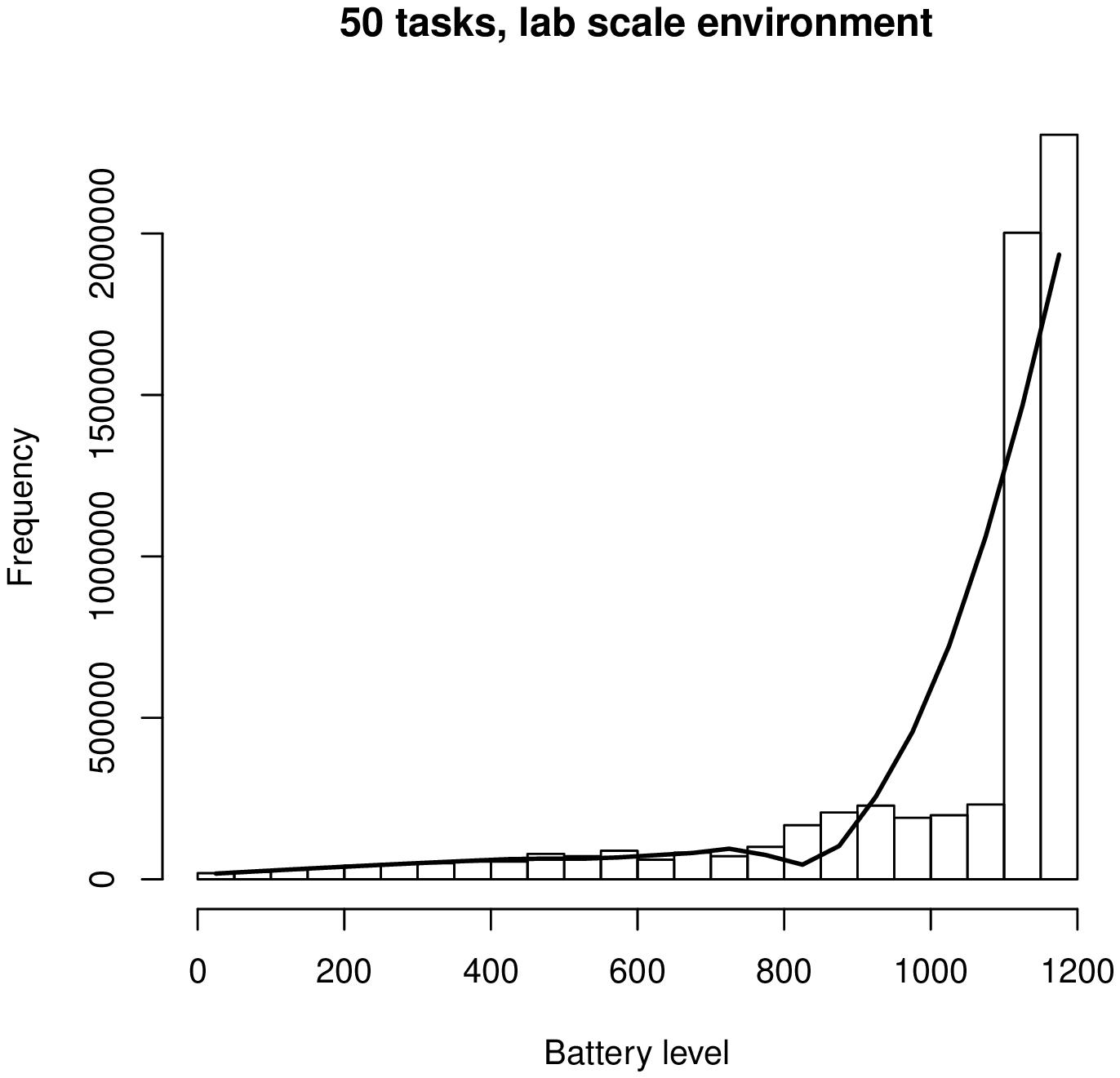}
  \caption{}\label{fig:battery_50_1}
  \end{subfigure}
  \begin{subfigure}{0.3\textwidth}
  \includegraphics[width=\linewidth]{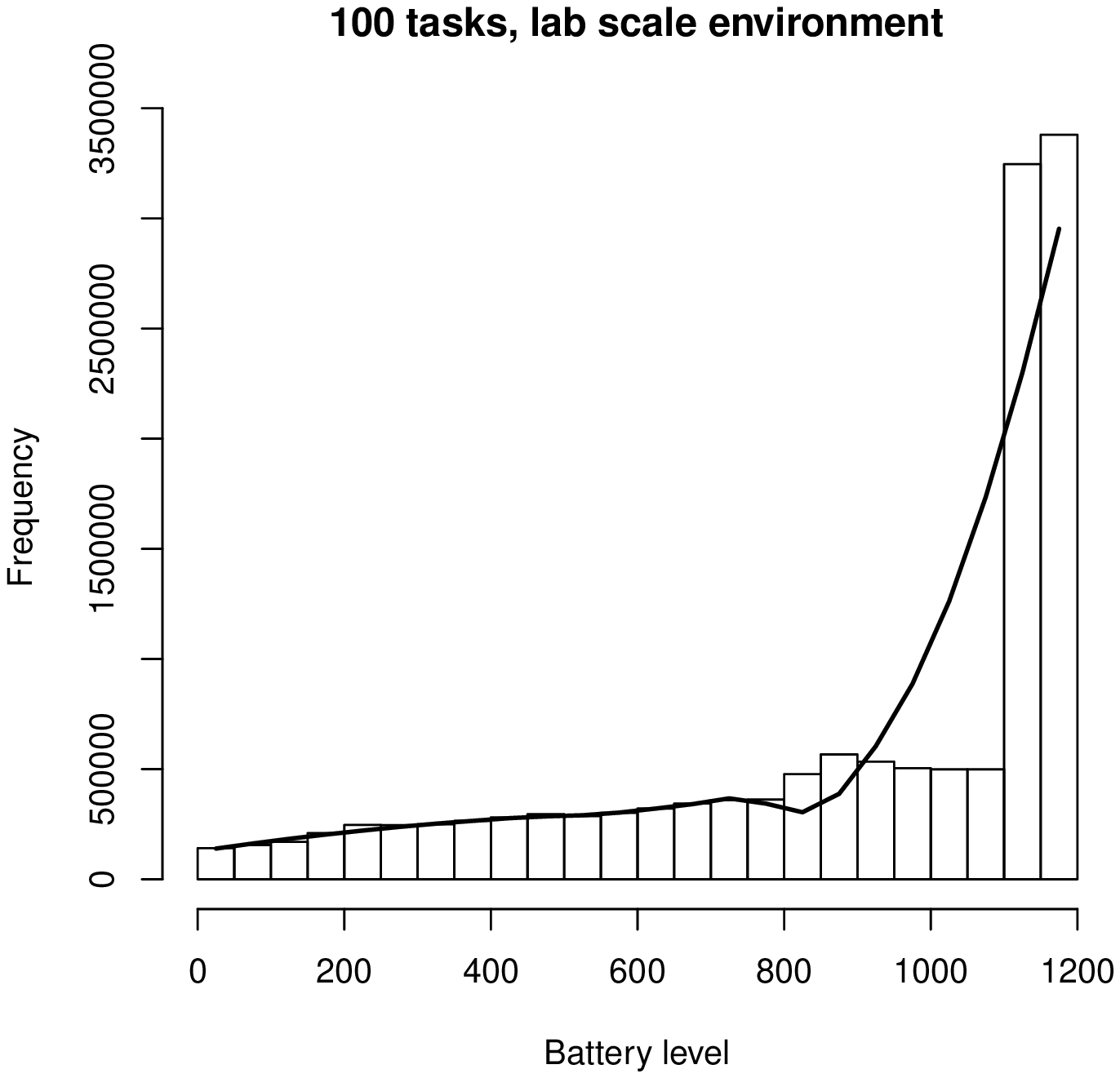}
  \caption{}\label{fig:battery_100_1}
  \end{subfigure}
  \begin{subfigure}{0.3\textwidth}
  \includegraphics[width=\linewidth]{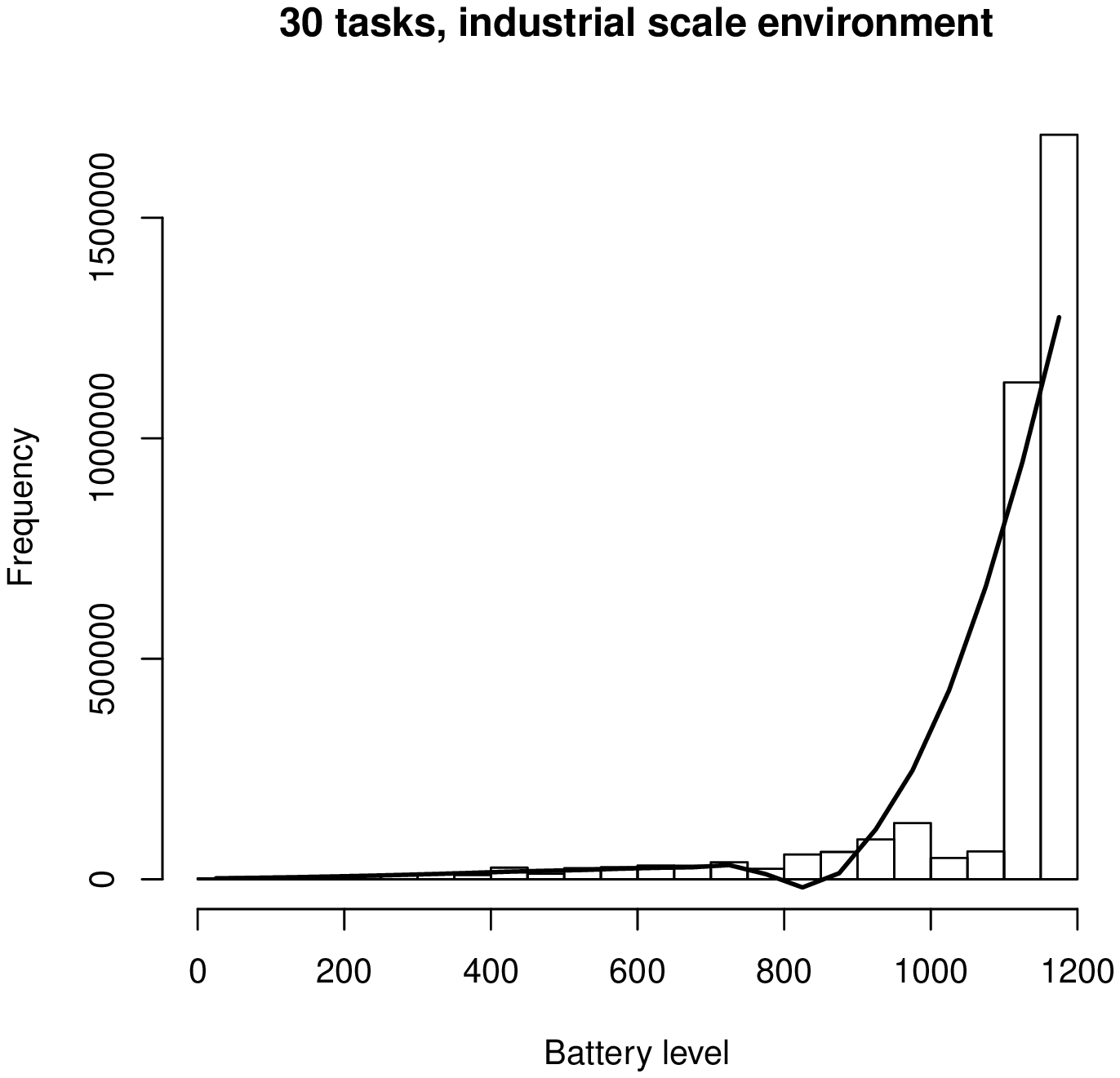}
  \caption{}\label{fig:battery_30_8}
  \end{subfigure}
  \begin{subfigure}{0.3\textwidth}
  \includegraphics[width=\linewidth]{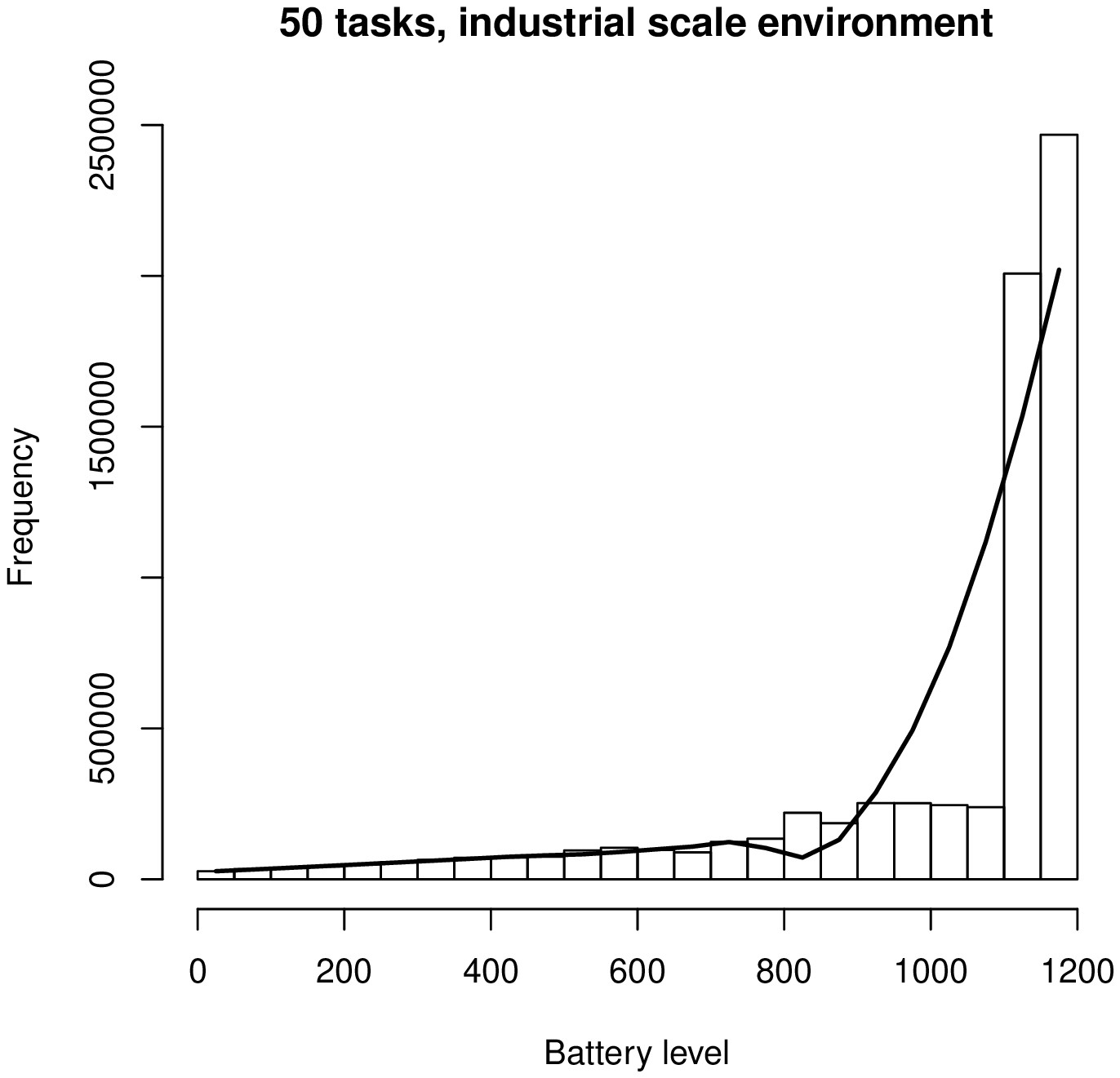}
  \caption{}\label{fig:battery_50_8}
  \end{subfigure}
  \begin{subfigure}{0.3\textwidth}
  \includegraphics[width=\linewidth]{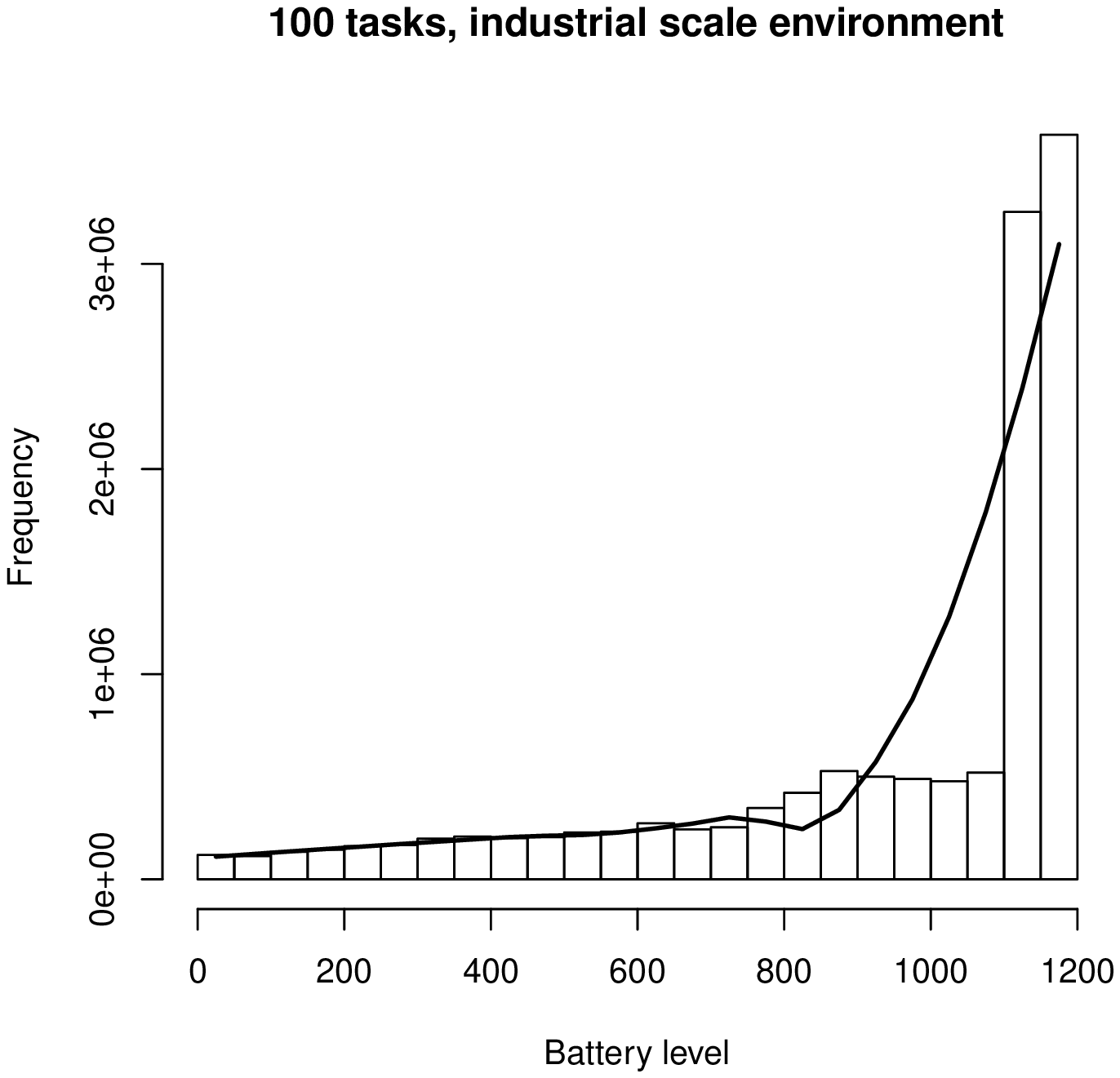}
  \caption{}\label{fig:battery_100_8}
  \end{subfigure}
  \caption{Battery level of the UAVs throughout the operations}
  \label{fig:battery}
\end{figure}

In alignment with the restful characteristic of Algorithm \ref{algo:algo_fragment}, the overall battery level is desired to be preserved throughout the operations. The battery level of each UAV after each task execution from every feasible schedule constructed during the search (iterations) is recorded, and the cumulative data is depicted through histograms in Figure \ref{fig:battery}. It is portrayed that the battery level of the UAVs has the tendency of being in a sufficiently preserved level. This condition applies for both lab scale and industrial scale environment, which shows the scalability of the proposed methodology.

\begin{figure}[H]
  \centering
  \begin{subfigure}{0.3\textwidth}
  \includegraphics[trim=10 15 50 0, clip, width=\linewidth]{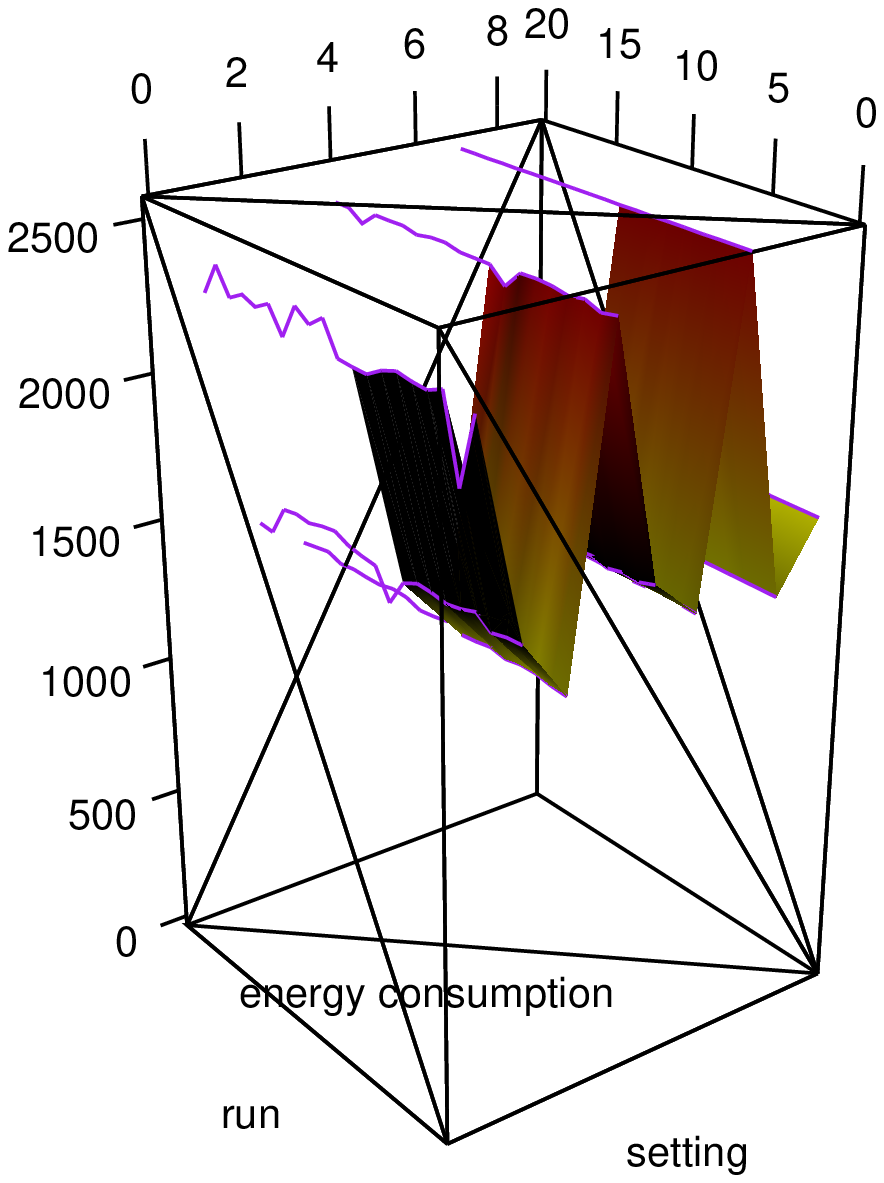}
  \caption{30 tasks, lab scale} \label{fig:result_30_1}
  \end{subfigure}
  \begin{subfigure}{0.3\textwidth}
  \includegraphics[trim=0 55 30 5, clip, width=\linewidth]{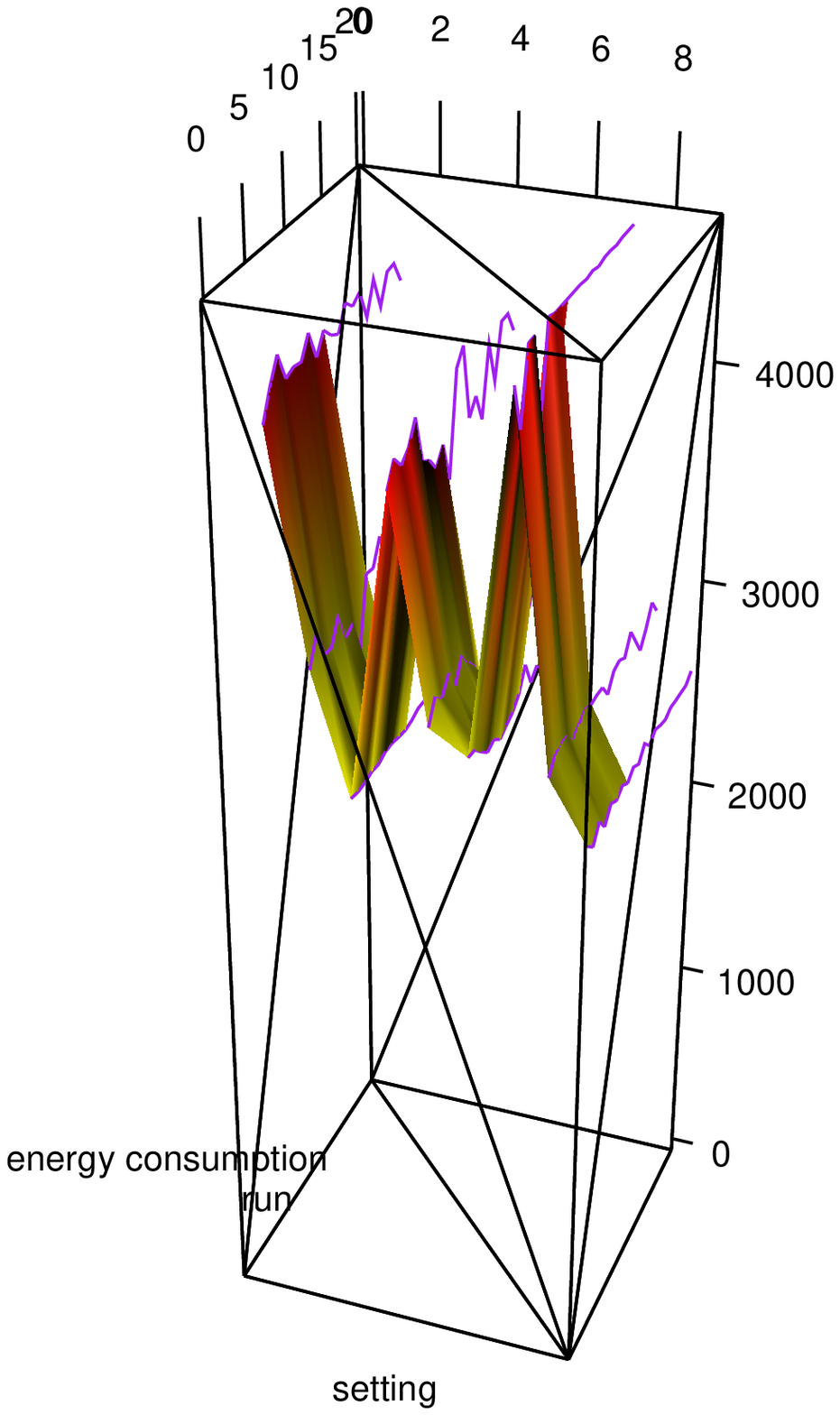}
  \caption{50 tasks, lab scale} \label{fig:result_50_1}
  \end{subfigure}
  \begin{subfigure}{0.3\textwidth}
  \includegraphics[trim=10 150 40 0, clip, width=\linewidth]{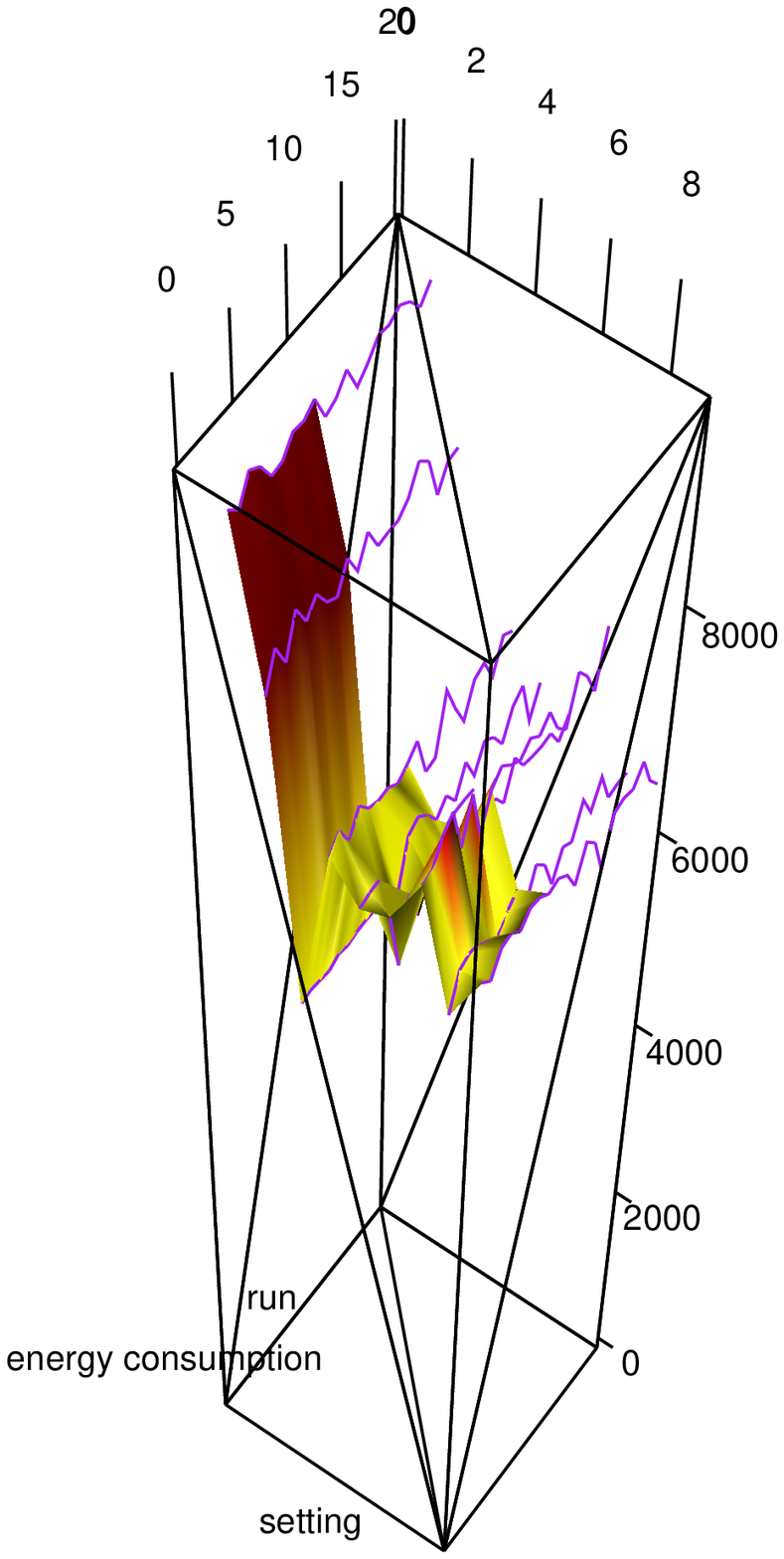}
  \caption{100 tasks, lab scale} \label{fig:result_100_1}
  \end{subfigure}
  \begin{subfigure}{0.3\textwidth}
  \includegraphics[trim=50 5 70 0, clip, width=\linewidth]{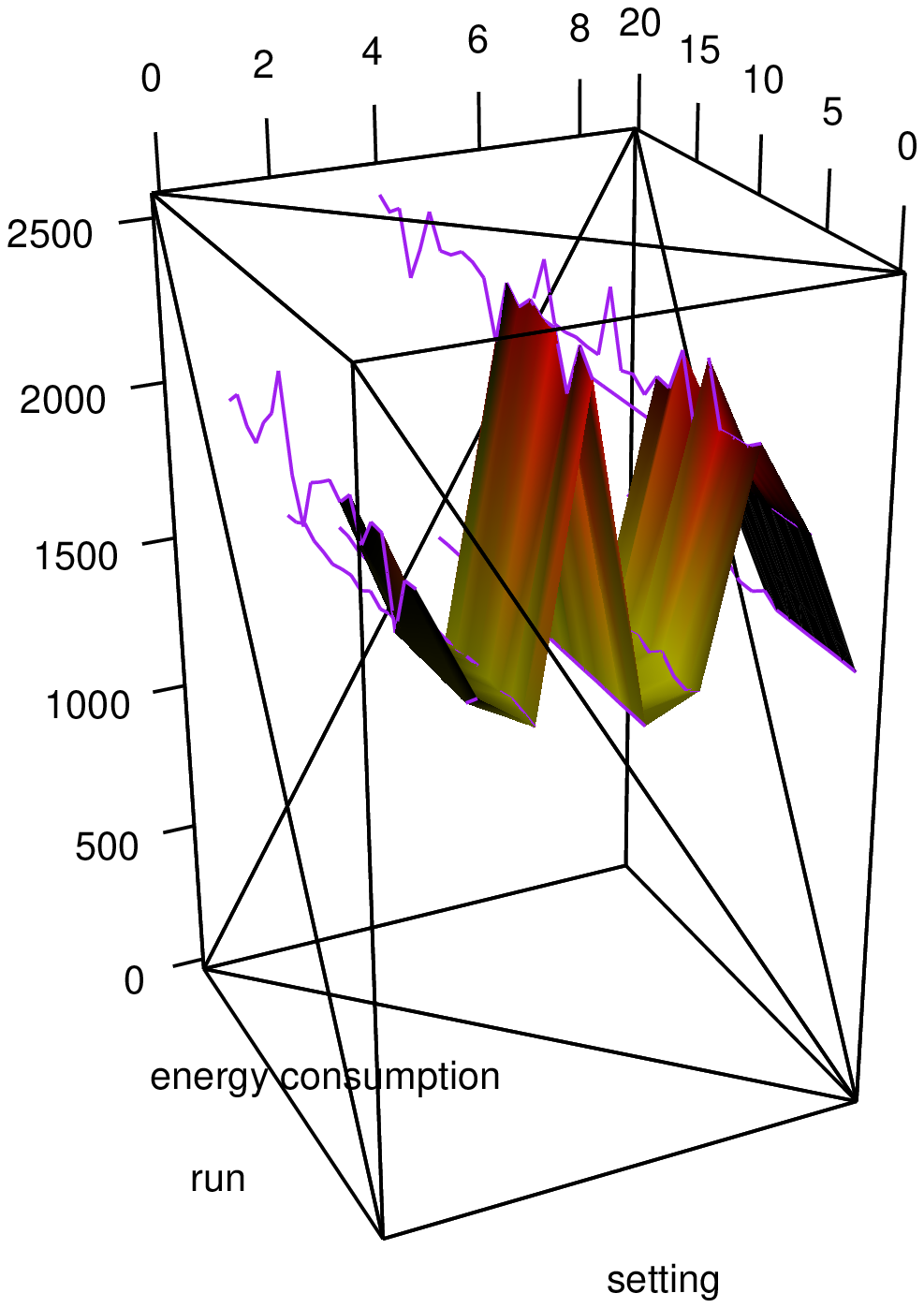}
  \caption{30 tasks, industrial scale} \label{fig:result_30_8}
  \end{subfigure}
  \begin{subfigure}{0.3\textwidth}
  \includegraphics[trim=20 30 50 0, clip, width=\linewidth]{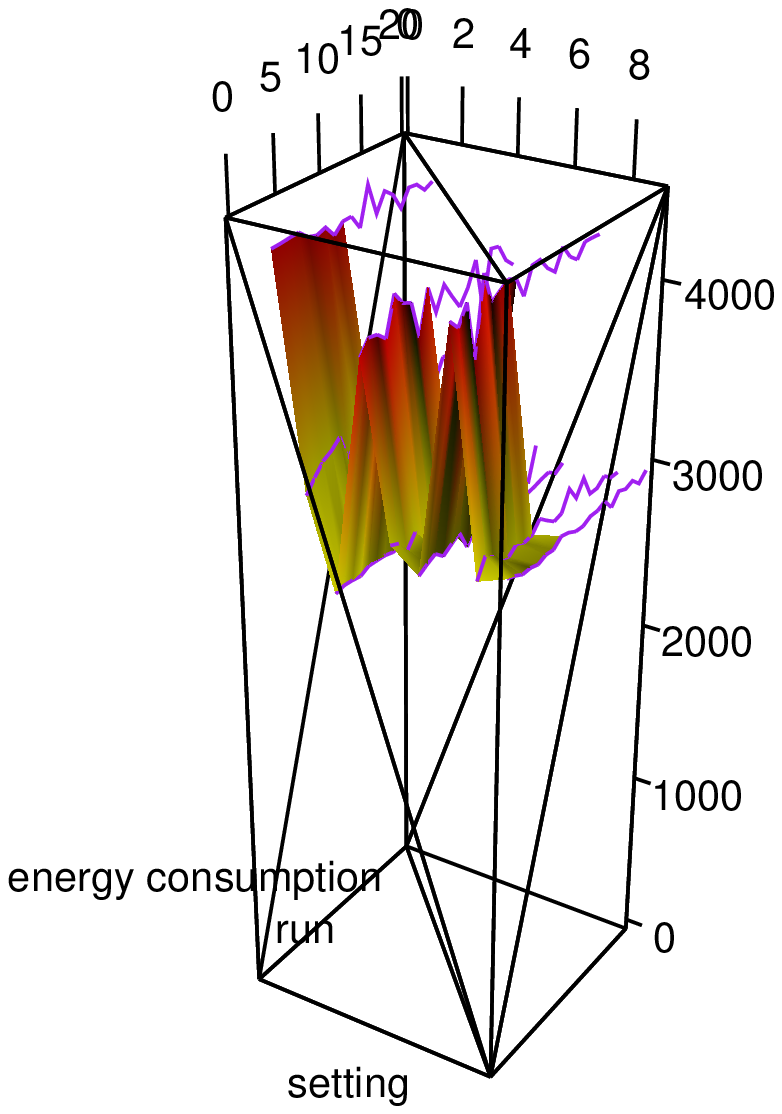}
  \caption{50 tasks, industrial scale} \label{fig:result_50_8}
  \end{subfigure}
  \begin{subfigure}{0.3\textwidth}
  \includegraphics[trim=10 195 35 75, clip, width=\linewidth]{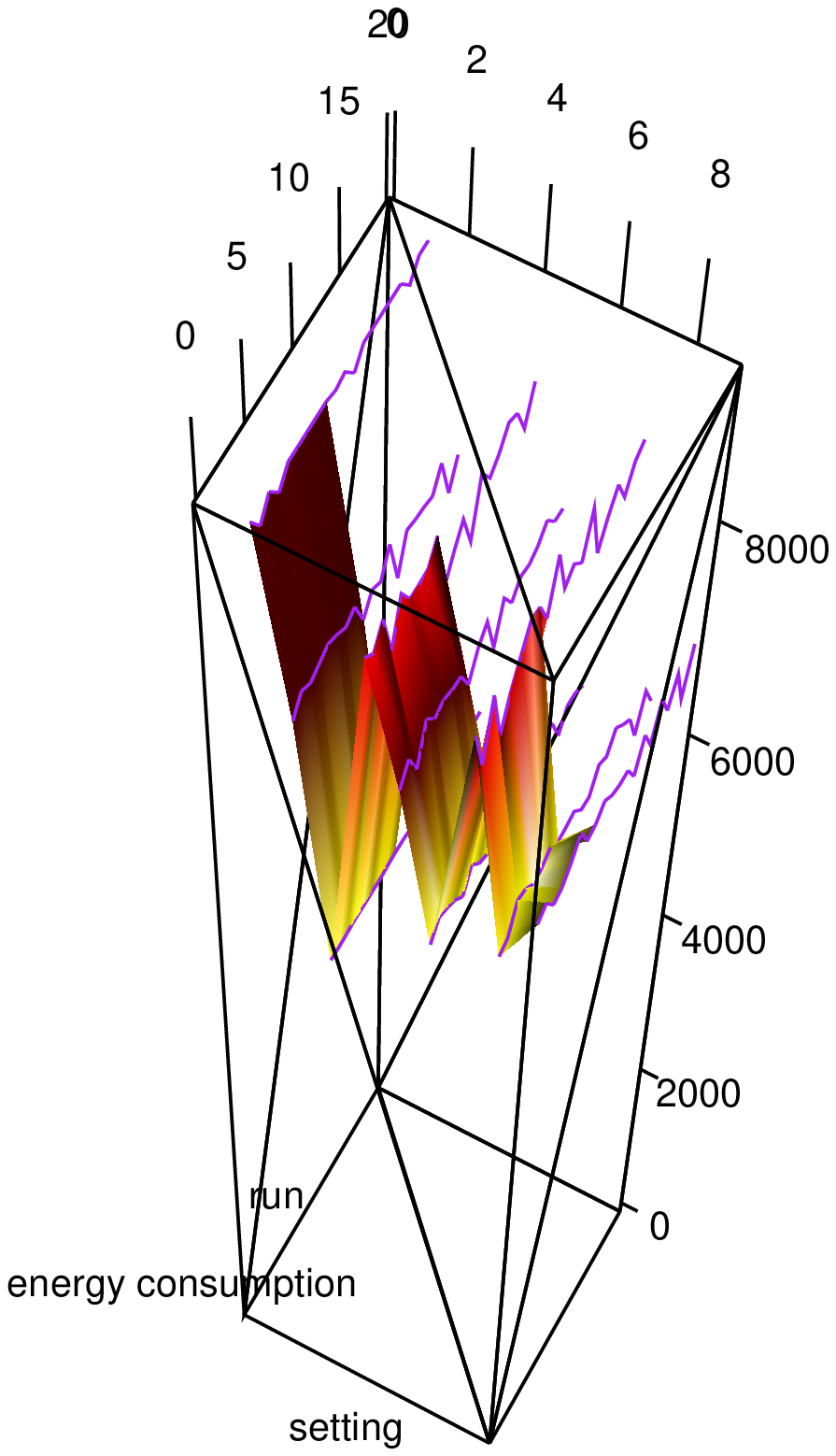}
  \caption{100 tasks, industrial scale} \label{fig:result_100_8}
  \end{subfigure}
  \caption{Battery consumption on UAV operations with 30, 50, and 100 tasks in lab scale and industrial scale indoor environments}
  \label{fig:results}
\end{figure}

 The set of learning coefficients used for the PSO algorithm is $c_1$=1, $c_2$=2, with 40 particles of initial swarm, 40 iterations, and a stop criteria of 10 no-improvement iterations \textemdash this is inspired by the pilot study conducted in \cite{khosiawan2016task}. In this study, the experiment purpose is to stress-test the methodology. Hence, the solution space may get vast in the used datasets, stressing exploration rooms not only for PSO, but also for RTAA. Each sub-figure in Figure \ref{fig:results} represents 180 attempts, which correspond to a particular geographical scale of the environment and number of tasks. There are 9 settings for a particular set of geographical scale and number of tasks, depicted in Table \ref{table:result_analysis_settings}. The battery consumption peaks are at setting 1, 4, and 7 for all sub-figures, where the slack time is 300. This behavior is caused by the situation where there is less time in between task executions, where UAVs will be required to hover and wait; since recharging at the recharge station within a very short period will be insignificant or even waste the battery more (due to the flights from and to the recharge station).
 In addition, it is depicted that the exposure of Algorithm \ref{algo:algo_prefUAVs} in Algorithm \ref{algo:algo_fragment_alap}\mytilde \ref{algo:algo_fragment_asap} (and Algorithm \ref{algo:algo_fragment} as a whole) may reduce the battery consumption in the produced schedule through the existence of precedence relationships.
 The results of the total battery consumption optimization in 1080 scheduling attempts are depicted in Figure \ref{fig:results}, and the numerical data of average battery consumption is presented in Table \ref{table:avg_batt_consum}.

\begin{table} [htp]
\caption{Result analysis settings}
\begin{center}
    \begin{tabular}{ c c c }
    \hline\noalign{\smallskip}
    Setting & Predecessor mean & Slack time mean (s)\\
    \noalign{\smallskip}\hline\noalign{\smallskip}
    1 & 0 & 300\\
    2 & 0 & 600\\
    3 & 0 & 1200\\
    4 & 1 & 300\\
    5 & 1 & 600\\
    6 & 1 & 1200\\
    7 & 2 & 300\\
    8 & 2 & 600\\
    9 & 2 & 1200\\ \hline
    \end{tabular}
\end{center}
\label{table:result_analysis_settings}
\end{table}

\begin{table}[htp]
\caption{Average battery consumption of the UAV operations in regard to the 54 benchmark datasets}
\centering
\resizebox{\textwidth}{!}{\begin{tabular}{l|r|r|ccc}
  \hline
   \multirow{15}{*}{Geographical distance - lab. scale (1)} & \multirow{5}{*}{Predecessor - none (0)} & \multirow{2}{*}{Number of tasks} & \multicolumn{3}{|c}{Slack time} \\\cline{4-6}
   & & & low (300) & medium (600) & high (1200) \\\cline{3-6}
   & & low - 30 & 2256.75 & 1490.4 & 1298.6 \\\cline{3-3}
   & & medium - 50 & 4230.4 & 3171.2 & 2531.4 \\\cline{3-3}
   & & high - 100 & 9305.05 & 8249.55 & 5387.55 \\\cline{2-6}
   & \multirow{5}{*}{Precedents - rare (1)} & \multirow{2}{*}{Task Size} & \multicolumn{3}{|c}{Slack time} \\\cline{4-6}
   & & & low (300) & medium (600) & high (1200) \\\cline{3-6}
   & & low - 30 & 2438.1 & 1547.65 & 1421.8 \\\cline{3-3}
   & & medium - 50 & 3957.65 & 2980.75 & 2783.05 \\\cline{3-3}
   & & high - 100 & 6953.55 & 6693.65 & 6600.65 \\\cline{2-6}
   & \multirow{5}{*}{Precedents - frequent (2)} & \multirow{2}{*}{Task Size} & \multicolumn{3}{|c}{Slack time} \\\cline{4-6}
   & & & low (300) & medium (600) & high (1200) \\\cline{3-6}
   & & low - 30 & 2532.75 & 1381 & 1610.4 \\\cline{3-3}
   & & medium - 50 & 4502.45 & 2848.75 & 2540.35 \\\cline{3-3}
   & & high - 100 & 7316.45 & 6466.15 & 6698.9 \\
  \hline
  \multirow{15}{*}{Geographical distance - industrial scale (8)} & \multirow{5}{*}{Predecessor - none (0)} & \multirow{2}{*}{Number of tasks} & \multicolumn{3}{|c}{Slack time} \\\cline{4-6}
   & & & low (300) & medium (600) & high (1200) \\\cline{3-6}
   & & low - 30 & 1921.55 & 1524.35 & 1429.4 \\\cline{3-3}
   & & medium - 50 & 4298.55 & 3219.7 & 2578.3 \\\cline{3-3}
   & & high - 100 & 9097.8 & 7595.45 & 5257.05 \\\cline{2-6}
   & \multirow{5}{*}{Precedents - rare (1)} & \multirow{2}{*}{Task Size} & \multicolumn{3}{|c}{Slack time} \\\cline{4-6}
   & & & low (300) & medium (600) & high (1200) \\\cline{3-6}
   & & low - 30 & 2434.4 & 1350.85 & 1407.1 \\\cline{3-3}
   & & medium - 50 &  3963.1 & 2855.2 & 2831.4 \\\cline{3-3}
   & & high - 100 & 8305.9 & 7516.8 & 6071.5 \\\cline{2-6}
   & \multirow{5}{*}{Precedents - frequent (2)} & \multirow{2}{*}{Task Size} & \multicolumn{3}{|c}{Slack time} \\\cline{4-6}
   & & & low (300) & medium (600) & high (1200) \\\cline{3-6}
   & & low - 30 & 2107.95 & 1825 & 1375.95 \\\cline{3-3}
   & & medium - 50 & 4173.4 & 2922.4 & 2922.05 \\\cline{3-3}
   & & high - 100 & 8276.55 & 6402.4 & 6766.4 \\
  \hline
\end{tabular}}
\label{table:avg_batt_consum}
\end{table}

\begin{figure}[H]
  \centering
  \begin{subfigure}{0.3\textwidth}
  \includegraphics[width=\linewidth]{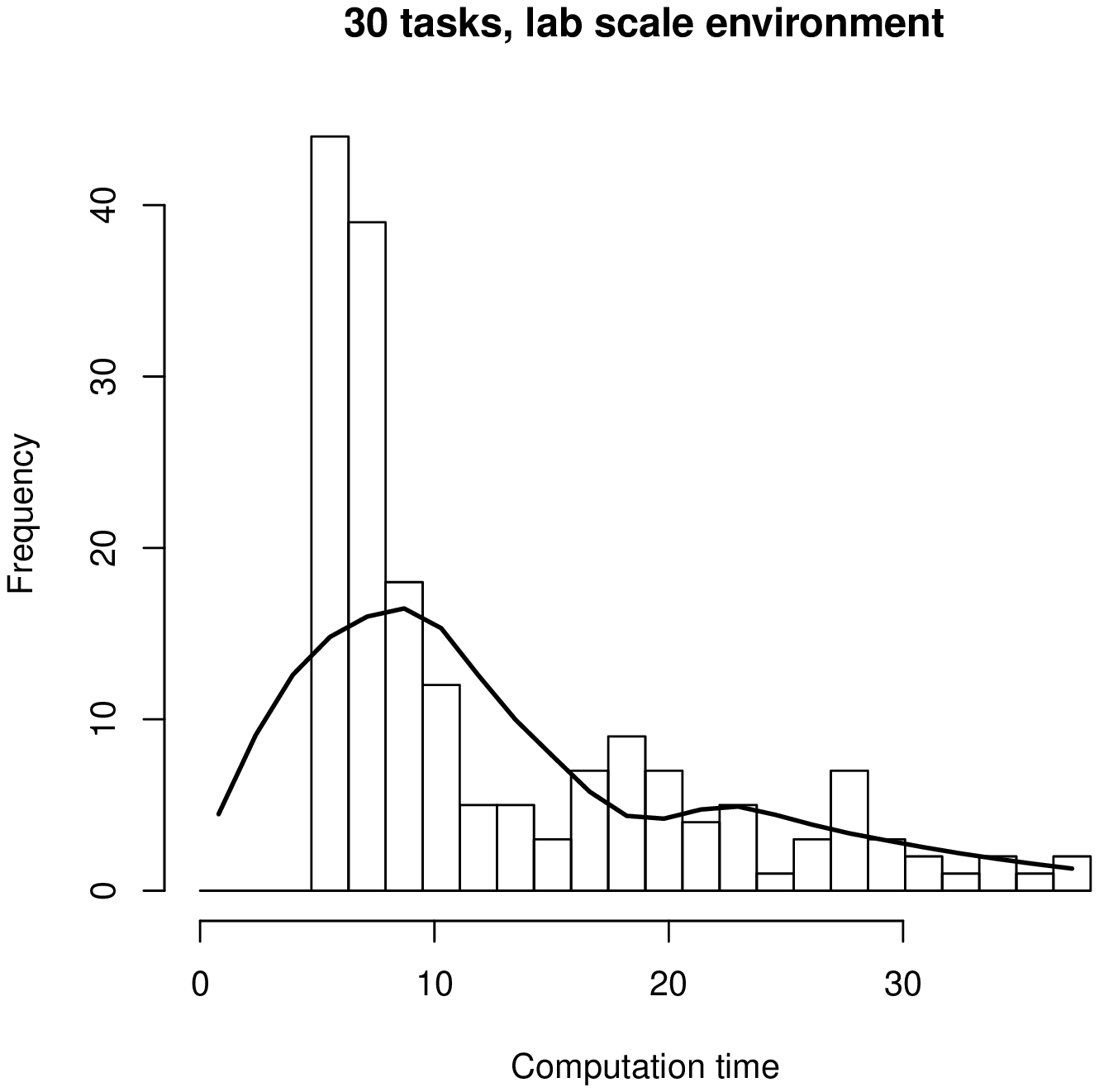}
  \caption{30 tasks, lab scale} \label{fig:comptime_30_1}
  \end{subfigure}
  \begin{subfigure}{0.3\textwidth}
  \includegraphics[width=\linewidth]{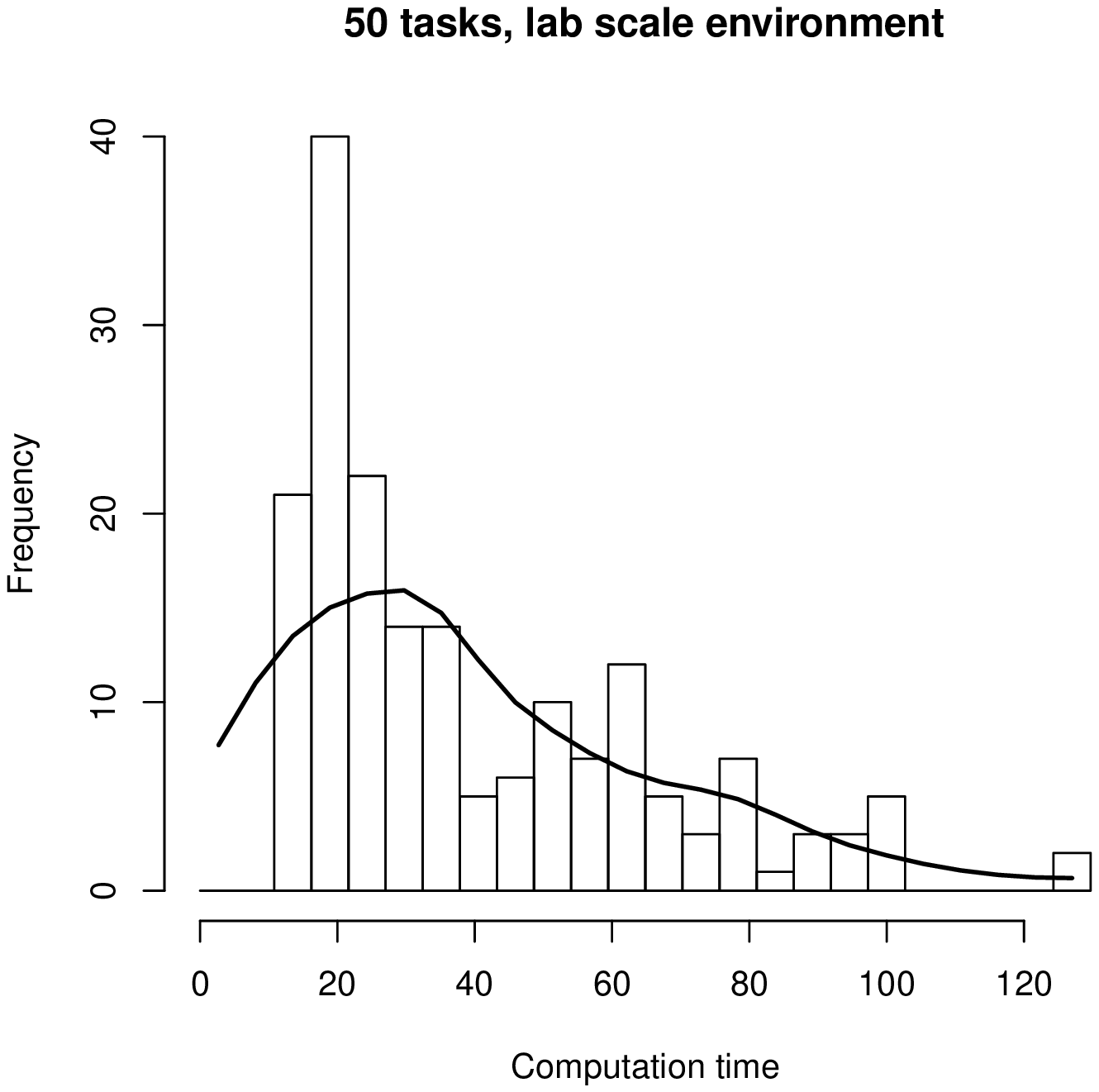}
  \caption{50 tasks, lab scale} \label{fig:comptime_50_1}
  \end{subfigure}
  \begin{subfigure}{0.3\textwidth}
  \includegraphics[width=\linewidth]{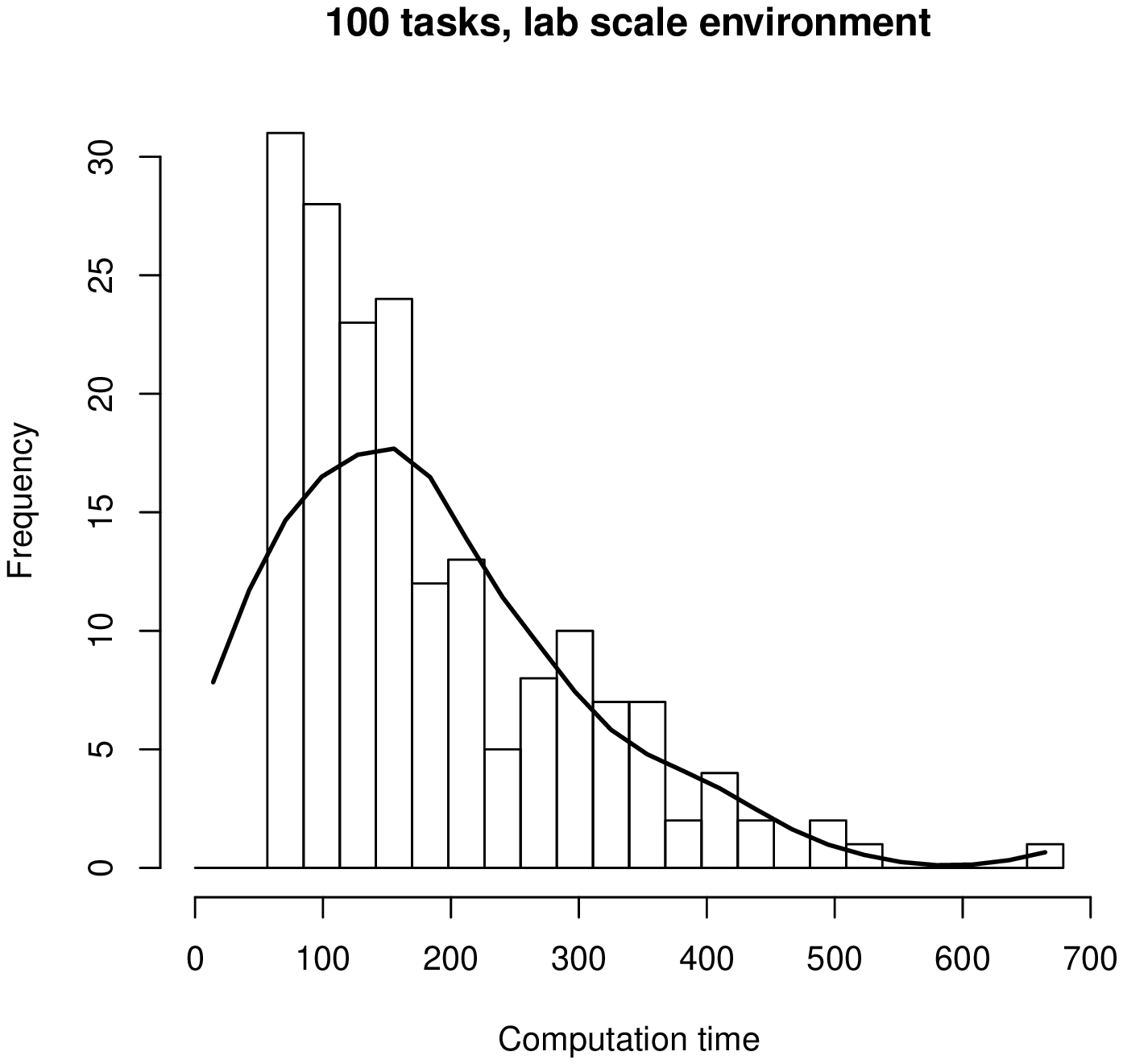}
  \caption{100 tasks, lab scale} \label{fig:comptime_100_1}
  \end{subfigure}
  \begin{subfigure}{0.3\textwidth}
  \includegraphics[width=\linewidth]{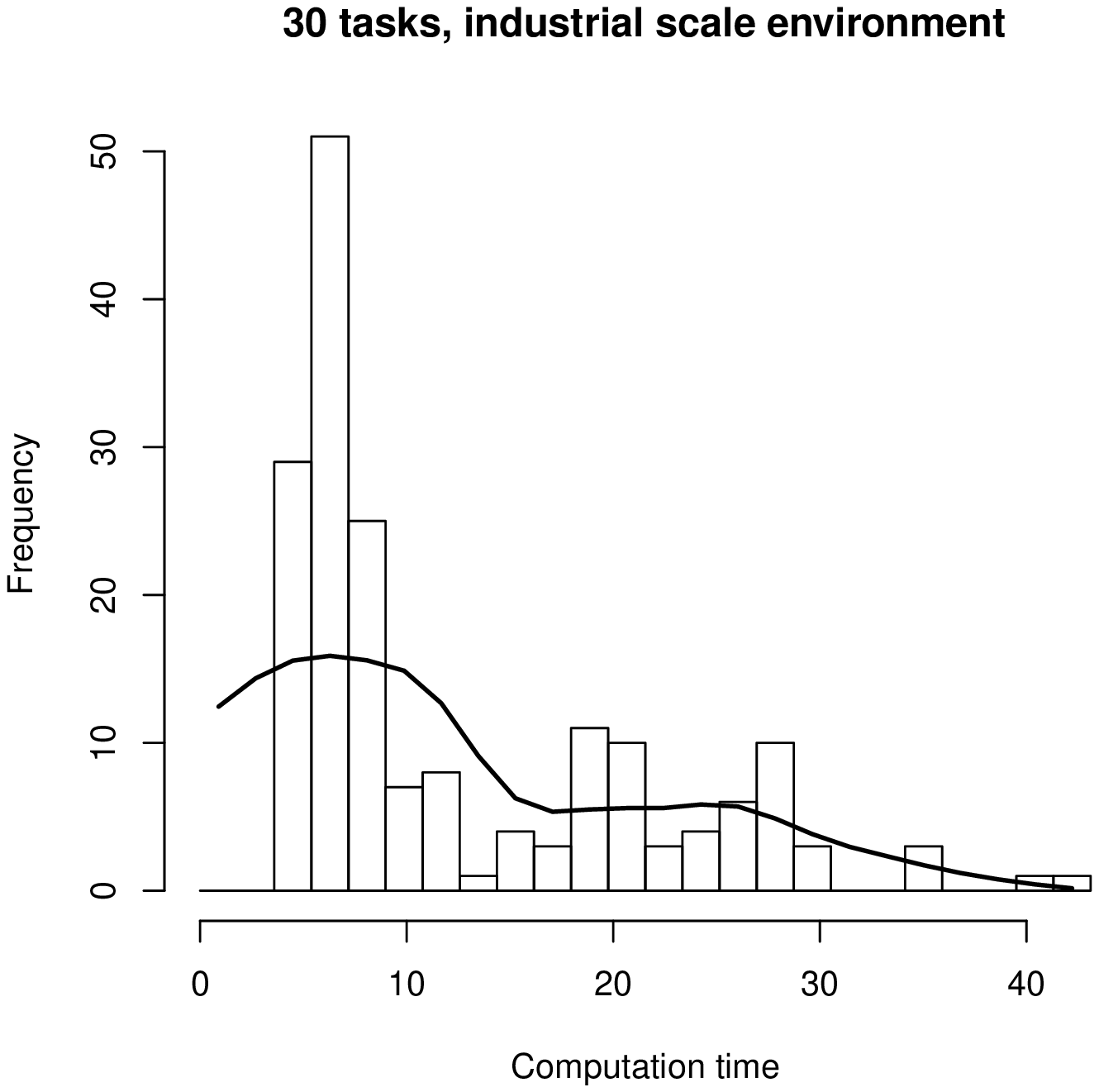}
  \caption{30 tasks, industrial scale} \label{fig:comptime_30_8}
  \end{subfigure}
  \begin{subfigure}{0.3\textwidth}
  \includegraphics[width=\linewidth]{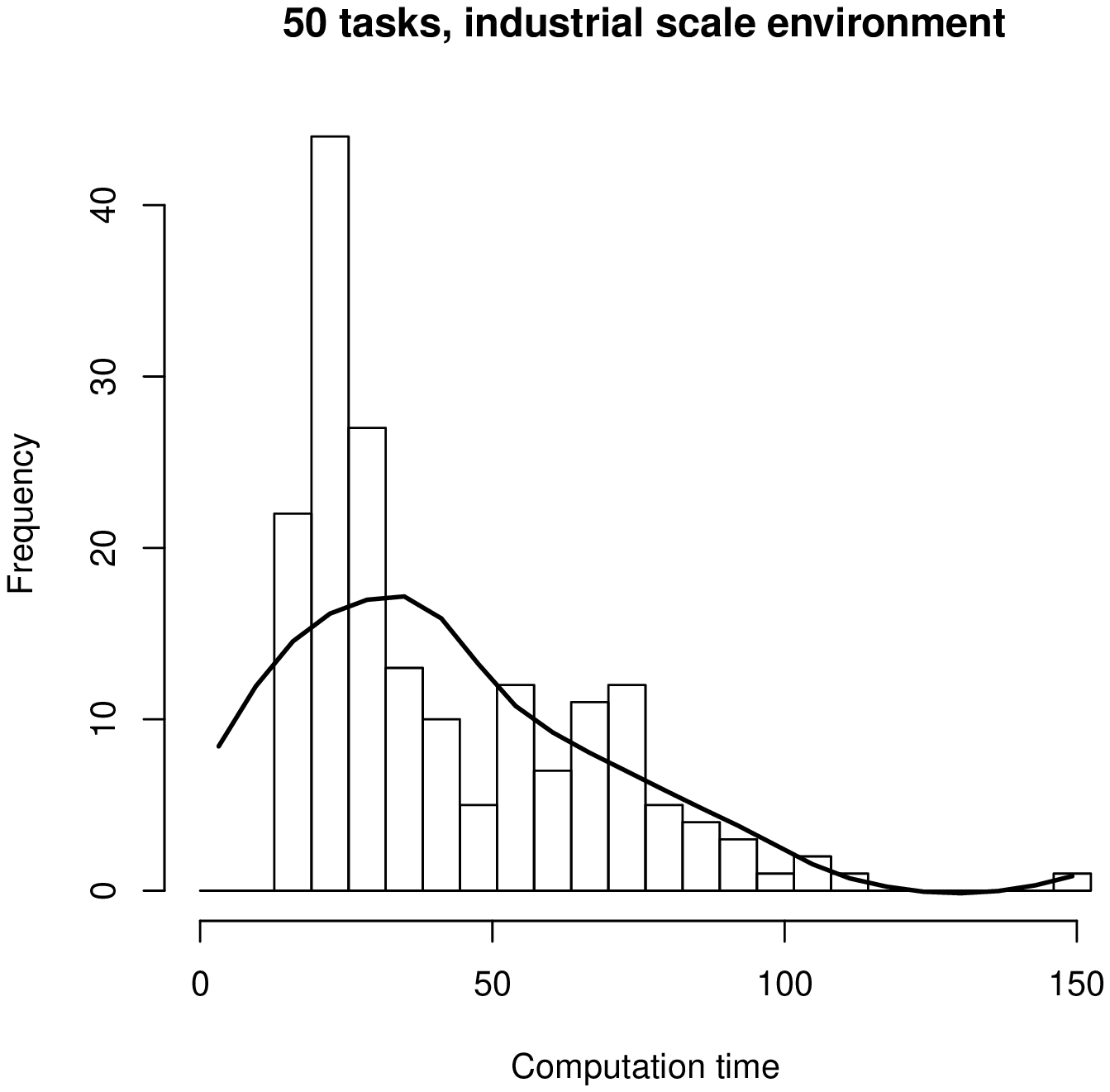}
  \caption{50 tasks, industrial scale} \label{fig:comptime_50_8}
  \end{subfigure}
  \begin{subfigure}{0.3\textwidth}
  \includegraphics[width=\linewidth]{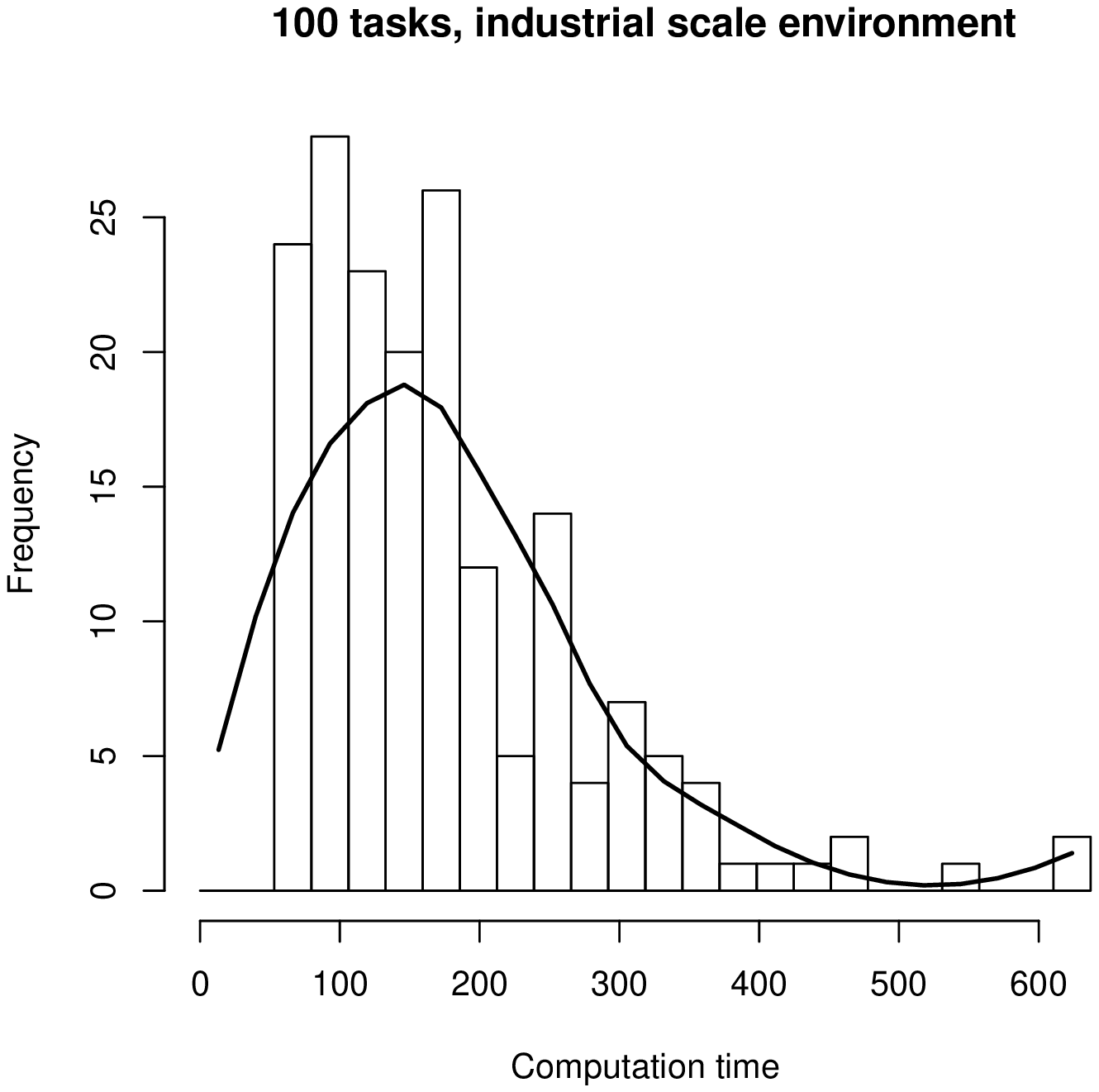}
  \caption{100 tasks, industrial scale} \label{fig:comptime_100_8}
  \end{subfigure}
  \caption{Computation time of scheduling attempt on UAV operations with 30, 50, and 100 tasks in lab scale and industrial scale indoor environments}
  \label{fig:comptime}
\end{figure}

In regard to the computation time, it tends to converge in the reasonable computation time area. The computation time of every 20 scheduling attempts of a particular dataset is recorded, and the respective overview for the benchmark datasets is depicted in Figure \ref{fig:comptime}. In Table \ref{table:avg_comp_time}, the average computation time of every dataset is presented.

\begin{table}[htp]
\caption{Average computation time (s) of scheduling attempts in regard to the 54 benchmark datasets}
\centering
\resizebox{\textwidth}{!}{\begin{tabular}{l|r|r|ccc}
  \hline
   \multirow{15}{*}{Geographical distance - lab. scale (1)} & \multirow{5}{*}{Predecessor - none (0)} & \multirow{2}{*}{Number of tasks} & \multicolumn{3}{|c}{Slack time} \\\cline{4-6}
   & & & low (300) & medium (600) & high (1200) \\\cline{3-6}
   & & low - 30 & 19.96 & 25.50 & 22.47 \\\cline{3-3}
   & & medium - 50 & 58.16 & 78.56 & 74.83 \\\cline{3-3}
   & & high - 100 & 245.02 & 297.45 & 380.85 \\\cline{2-6}
   & \multirow{5}{*}{Precedents - rare (1)} & \multirow{2}{*}{Task Size} & \multicolumn{3}{|c}{Slack time} \\\cline{4-6}
   & & & low (300) & medium (600) & high (1200) \\\cline{3-6}
   & & low - 30 & 8.69 & 8.12 & 7.68 \\\cline{3-3}
   & & medium - 50 & 24.59 & 26.63 & 36.51 \\\cline{3-3}
   & & high - 100 & 144.69 & 133.03 & 119.83 \\\cline{2-6}
   & \multirow{5}{*}{Precedents - frequent (2)} & \multirow{2}{*}{Task Size} & \multicolumn{3}{|c}{Slack time} \\\cline{4-6}
   & & & low (300) & medium (600) & high (1200) \\\cline{3-6}
   & & low - 30 & 7.06 & 6.01 & 7.21 \\\cline{3-3}
   & & medium - 50 & 19.04 & 19.73 & 19.05 \\\cline{3-3}
   & & high - 100 & 109.60 & 112.59 & 121.15 \\
  \hline
  \multirow{15}{*}{Geographical distance - industrial scale (8)} & \multirow{5}{*}{Predecessor - none (0)} & \multirow{2}{*}{Number of tasks} & \multicolumn{3}{|c}{Slack time} \\\cline{4-6}
   & & & low (300) & medium (600) & high (1200) \\\cline{3-6}
   & & low - 30 & 24.14 & 20.51 & 26.02 \\\cline{3-3}
   & & medium - 50 & 63.04 & 74.14 & 77.47 \\\cline{3-3}
   & & high - 100 & 185.33 & 306.68 & 352.62 \\\cline{2-6}
   & \multirow{5}{*}{Precedents - rare (1)} & \multirow{2}{*}{Task Size} & \multicolumn{3}{|c}{Slack time} \\\cline{4-6}
   & & & low (300) & medium (600) & high (1200) \\\cline{3-6}
   & & low - 30 & 7.50 & 6.77 & 8.60 \\\cline{3-3}
   & & medium - 50 &  28.06 & 27.95 & 27.27 \\\cline{3-3}
   & & high - 100 & 151.92 & 112.47 & 160.97 \\\cline{2-6}
   & \multirow{5}{*}{Precedents - frequent (2)} & \multirow{2}{*}{Task Size} & \multicolumn{3}{|c}{Slack time} \\\cline{4-6}
   & & & low (300) & medium (600) & high (1200) \\\cline{3-6}
   & & low - 30 & 6.74 & 5.21 & 6.83 \\\cline{3-3}
   & & medium - 50 & 26.46 & 23.03 & 26.01 \\\cline{3-3}
   & & high - 100 & 109.90 & 84.19 & 116.52 \\
  \hline
\end{tabular}}
\label{table:avg_comp_time}
\end{table}

\section{Conclusion}
\label{sec:sec_conclusion}
UAV scheduling problem in indoor environment has been emerging with various challenges to solve. In this paper, a study on safe UAV scheduling is conducted, where tasks with time windows are performed by multiple UAVs in indoor environment. A heuristic approach based on RTAA is proposed. Furthermore, an incorporation of PSO with RTAA is performed, where two additional priority rules are proposed apart from the ones utilized in the pilot study \cite{khosiawan2016task}. The study involves 54 benchmark datasets stressing on different aspects: geographical distance, number of tasks, number of predecessors, and slack time. From the analysis of the results, it can be deduced that the proposed methodology is very effective for producing a restful schedule in a reasonable amount of time. This depicts a safe UAV scheduling, where the battery level is shown to be sufficiently preserved throughout the operations. On top of that, RTAA (specifically Algorithm \ref{algo:algo_prefUAVs},\ref{algo:algo_fragment_alap}, and \ref{algo:algo_fragment_asap}) which is incorporated with PSO generates a drive for the search to obtain minimum total battery consumption. This methodology is also shown to be scalable for both lab scale and industrial scale indoor environments.

\section*{Conflicts of Interest}
The authors declare that they have no conflicts of interest.

\section*{Acknowledgement}
This work has partly been supported by Innovation Fund Denmark under project UAWorld; grant agreement number 9-2014-3.


\bibliography{citations}{}
\bibliographystyle{plain}

\end{document}